\documentclass[journal]{IEEEtran}
\usepackage[table]{xcolor}
\ifCLASSINFOpdf
   \usepackage[pdftex]{graphicx}
\else
\fi
\usepackage[ruled, lined, longend, linesnumbered]{algorithm2e}

\usepackage{float}
\usepackage{subfig}
\usepackage{fixltx2e}
\usepackage{tikz,xcolor,hyperref}
\definecolor{lime}{HTML}{A6CE39}
\DeclareRobustCommand{\orcidicon}{%
	\begin{tikzpicture}
	\draw[lime, fill=lime] (0,0) 
	circle [radius=0.16] 
	node[white] {{\fontfamily{qag}\selectfont \tiny ID}};
	\draw[white, fill=white] (-0.0625,0.095) 
	circle [radius=0.007];
	\end{tikzpicture}
	\hspace{-2mm}
}
\foreach \x in {A, ..., Z}{%
	\expandafter\xdef\csname orcid\x\endcsname{\noexpand\href{https://orcid.org/\csname orcidauthor\x\endcsname}{\noexpand\orcidicon}}
}

\begin{document}
\title{Evolving Character-level Convolutional Neural Networks for Text Classification}
\author{Trevor~Londt\orcidA{},~\IEEEmembership{Member,~IEEE,}
        Xiaoying~Gao\orcidB{},~\IEEEmembership{Member,~IEEE,}
        Bing~Xue\orcidC{},~\IEEEmembership{Member,~IEEE}
        and~Peter~Andreae\orcidD{},~\IEEEmembership{Member,~IEEE}}
\maketitle
\begin{abstract}
Character-level convolutional neural networks (char-CNN) require no knowledge of the semantic or syntactic structure of the language they classify. This property simplifies its implementation but reduces its classification accuracy. Increasing the depth of char-CNN architectures does not result in breakthrough accuracy improvements. Research has not established which char-CNN architectures are optimal for text classification tasks. Manually designing and training char-CNNs is an iterative and time-consuming process that requires expert domain knowledge. Evolutionary deep learning (EDL) techniques, including surrogate-based versions, have demonstrated success in automatically searching for performant CNN architectures for image analysis tasks. Researchers have not applied EDL techniques to search the architecture space of char-CNNs for text classification tasks. This article demonstrates the first work in evolving char-CNN architectures using a novel EDL algorithm based on genetic programming, an indirect encoding and surrogate models, to search for performant char-CNN architectures automatically. The algorithm is evaluated on eight text classification datasets and benchmarked against five manually designed CNN architectures and one long short-term memory (LSTM) architecture. Experiment results indicate that the algorithm can evolve architectures that outperform the LSTM in terms of classification accuracy and five of the manually designed CNN architectures in terms of classification accuracy and parameter count.
\end{abstract}
\begin{IEEEkeywords}
Character-level convolutional neural network, evolutionary deep learning, genetic programming,  text classification.
\end{IEEEkeywords}
\IEEEpeerreviewmaketitle
\section{Introduction}
DEEP learning \cite{Lecun2015} is a modern machine learning technique based on artificial neural networks. The field of natural language processing (NLP) has significantly benefited from the use of deep learning techniques in recent years \cite{Collobert2011a}\cite{devlin2018bert}\cite{Zhang2015}\cite{Conneau2017}\cite{Yang2019}\cite{peters-etal-2018-deep}. There are three prevalent deep learning architectures concerned with  NLP tasks: long-short term memory (LSTM) \cite{LSTM} 
, transformer networks \cite{NIPS2017_7181} and convolutional neural networks (CNNs) \cite{Lecun1998}. LSTMs exhibit relatively slow inference speeds and are less performant than transformers and CNNs with regards to text classification accuracy \cite{Zhang2015}. Transformers are a recent innovation and have shown significant successes in many NLP tasks \cite{devlin2018bert}\cite{Yang2019}\cite{peters-etal-2018-deep}. Their massive complexity with trainable parameters in the order of hundreds of millions presents critical \emph{experiment reproducibility} challenges to researchers. State-of-the-art transformers are difficult to reproduce in lab conditions as they have a high training cost in monetary terms. There are only a limited number of pre-trained transformer models available for different languages.
\par
CNNs have demonstrated excellent success in text classification tasks \cite{Zhang2015}\cite{Conneau2017}\cite{Zhang2013}\cite{Kim2015}\cite{Johnson2016}. There are two paradigms available when using CNNs for text classification tasks, namely: world-level (word-CNN) \cite{Kim} and character-level CNNs \cite{Zhang2015}.
\par
Word-level approaches are dependant on a word-model to represent the text. The reliance on a pre-trained word-model poses the potential problem of not having one available for a particular language. Training new word models is computationally time-consuming and costly. There is also the technical challenges of dealing with misspellings and words that may not exist in the word-model. The other paradigm is char-CNNs. No pre-trained language or word models are required. They also do not require a costly pre-processing step of the text data. In general, char-CNNs are not as accurate as word-level CNNs or transformers. Adding depth has not given the benefit of improved classification accuracy, as seen in image classification tasks. There is an open question in the research literature of what is the optimal architecture for char-CNNs. Little research has been performed to address these limitations. Deep learning is an iterative process requiring the tuning of many hyper-parameters and repeated experiments to test the efficacy of any potential architecture. It is a time consuming, costly and a tedious process that requires expert skills and domain knowledge. The task of finding optimal char-CNNs is an NP-hard problem.
\par
Evolutionary computation (EC) \cite{7339682} is a collection of search algorithms inspired by the principals of biological evolution, in particular the concept of \emph{survival of the fittest}. EC methods use a population of individuals (candidate solutions) to conduct a simultaneous search during a limited time frame to improve the optimisation of a specified objective function via the exchange of information between individuals in the population. The exchange of information is one of the key motivating factors of selecting EC methods for evolving char-CNNs in this work. There is the potential that this information exchange may reveal the essential characteristics of what makes a non-performant char-CNN into a performant one. EC methods are concerned with locating near-optimal solutions to NP-hard problems.
\par
Evolutionary deep learning (EDL) is the technique of using EC methods to search for candidate CNN architectures combined with the backpropagation algorithm to train any potential candidate network architecture. EDL has demonstrated success when searching for performant CNN architectures on image classification tasks \cite{sun1}\cite{8571181}\cite{8477735}. EDL has not been used to search for performant char-CNN architectures.
\par
Motivated by the success of applying EDL techniques in the image classification domain, we propose a novel surrogate-based EDL algorithm appropriate for searching the landscape of char-CNN architectures for the text classification domain. The proposed algorithm is based on genetic programming (GP) and an indirect encoding that is capable of representing novel char-CNN  architectures. The algorithm employs the use of surrogate models to significantly reduce the training time of the candidate char-CNNs during the evolutionary process.

In summary, the contributions of the proposed algorithm and work are:
\begin{enumerate}
  \item A fully automated approach to constructing, training and evaluating char-CNNs of variable length and complexity.
  \item A surrogate model approach that significantly reduces the computational time required to evolve performant char-CNNs.
  \item An expressive indirect encoding that ensures that all evolved candidate networks in the population are structurally valid and trainable networks, thereby preventing wasted computational power and time.
  \item Evidence that branching (width) in the early stages of a char-CNNs architecture may aid in improving classification performance.
  \item Evidence that the genealogy of an evolved char-CNN can provide insights into the architectural properties that aid in improving char-CNN performance. 
\end{enumerate}
\section{LITERATURE REVIEW}
\subsection{Background}
\subsubsection{Character-level Convolutional Neural Networks}
Kim et al. \cite{Kim} were the first to use a CNN with pre-trained word embeddings, word2vec \cite{Mikolov}, to perform sentence-level text classification. Their simple CNN architecture with one convolutional layer and a single max-pooling layer outperformed state-of-the-art traditional methods on four of the seven datasets tested. Filter widths of 3, 4 and 5 each having 100 channels were implemented. Their choice of activation function for non-linearity was the ReLU \cite{relu} function. It should be noted that their model is shallow. Notably, their baseline model with randomly initialised word embeddings performed poorly relative to all other models. This finding highlighted the importance of word2vec in their performance gains. Another interesting finding was that the use of dropout as a regularisation technique provided a 2-4\% performance increase across all datasets. Although their model achieved good performance, it should be noted that all the datasets used were relatively small in size. A limitation is that their work was only conducted across English datasets and has not been proven to work with other languages.
\par
Zhang et al. \cite{Zhang2015} were the first to conduct research on the use of char-CNNs for text classification. Their model had a modular design using back-propagation \cite{backprop} for gradient optimisation via a stochastic gradient descent \cite{8192502} algorithm. The main component of their design was a temporal convolutional module that computed a one-dimensional convolution. Max-pooling was implemented to allow their network to be deeper than six layers. ReLU \cite{relu} was used for non-linearity. The classifier section of the network was two fully connected layers. The text was encoded by converting each character in the sequence of text as a one-hot vector. The vector was created according to a lookup table consisting of a predetermined alphabet of length $m$. A maximum sentence length of 1014 characters was specified. A sequence of characters of length $j$ would have 1014 one-hot vectors, each of length $m$. Any characters beyond the maximum length were ignored. Their experiment was conducted over eight datasets. The datasets were constructed by the authors from large publicly available datasets and were chosen to represent different tasks and volume sizes. The datasets have now become the standard for testing char-CNNs. The major finding of their paper was that char-CNNs are an effective approach for text classification. It was shown that their model performed better on larger datasets than smaller datasets. According to their findings, traditional approaches performed well until the datasets approached the scale of millions of instances. Another interesting insight was that the choice of the alphabet made a significant difference in the accuracy performance. Zhang et al. \cite{Zhang2015} demonstrated the utility of char-CNNs. However, their model was not particularly deep when compared to CNNs used for image classification tasks.
\par
Conneau et al. \cite{Conneau2017} demonstrated the benefits of adding depth to a char-CNN with their \emph{very deep convolutional neural network} (VDCNN) model. Their model was built in a modular format where they used the concept of a \emph{convolutional block} stacked multiple times in sequence one after the other. Each convolutional block consisted of a convolutional layer followed by a temporal batch normalisation \cite{Ioffe2015c} layer and then a ReLu activation function. This sequence is repeated twice in each block. Implementing shortcut links, inspired by ResNet skip links \cite{resnet}, their model was able to be extended to a depth of 29 layers. Their model outperformed all current state-of-the-art models on the eight datasets introduced by Zhang et al. \cite{Zhang2015}. VDCNN demonstrated the advantage of adding depth to a char-CNN to increase performance. The caveat to their findings was that depth only increased performance up to a certain depth after which adding additional layers degraded the model's performance. Their deepest model reached 49 layers and had a reduced relative accuracy of approximately 3\% compared to the 29 layer model over the yelp dataset. The larger layer model was not tested over the other datasets.
\par
\par
Le et al. \cite{Le} conducted a study into the role of depth for both char-CNNs and word-CNNs for text classification. Motivated by the success of the state-of-the-art DenseNet \cite{Huang2016} model used for image classification tasks, Le et al.\cite{Le} implemented both a world-level and char-level DenseNet model. Their word-level DenseNet model used Word2vec for the word embeddings. The character-level DenseNet model used the same alphabet as in \cite{Zhang2015} and \cite{Conneau2017}. Both models were tested on only five of the original datasets in \cite{Zhang2015}. Both models performed comparatively similar to each other with the word-level DenseNet model being marginally better. Both models only slightly under-performed the shallower model in \cite{Johnson}. The main finding of their research is that adding depth to CNNs for text classification is still not a well-understood technique. Although there has been an increase in performance with depth, the increase has not been substantial. A second finding is that the state-of-the-art DenseNet model did not provide the same breakthrough improvements as seen in image classification tasks. The authors conclude that if a char-CNN is to be used then the model \emph{must} be deep. However, it is not yet known what architectures can further improve char-CNN performance to the level of word-CNNs performance.
\subsection{Related work}
The search for network architectures is currently an interesting and challenging research task. However, evolving char-CNNs for text classification is a nascent research topic and there is no research work directly related to evolving char-CNNs. However it is worth noting the work of Liang et al. \cite{Liang2019}. Their work presented an evolutionary-based framework named LEAF that simultaneously evolved network architectures and optimised hyperparameters. Their algorithm consisted of three components: an algorithm layer, a system layer and a problem-domain layer. The algorithm layer was responsible for evolving network topologies and hyperparameters. The system layer distributed the training of the networks across multiple cloud computing services. The algorithm and system layer cooperated to support the problem-domain layer, and the problem-domain layer performed hyperparameter tuning and architecture search. The algorithm layer was based on a cooperative co-evolution algorithm named CoDeepNEAT \cite{miikkulainen2017evolving}. A population of network architectures of minimal complexity and size were initially generated. The network architectures were all encoded as graphs. CoDeepNEAT was based on the NEAT \cite{Stanley2002} algorithm, where a mutation operation adds new nodes or connections to the network. The alignment of parent chromosomes facilitated the crossover operation according to historical markings placed on the genes of the chromosomes during the evolutionary process. This approach allowed segments of the network to be crossed over and remain a valid network structure. CoDeepNEAT differs from NEAT in that instead of nodes representing neurons; layers are represented instead. Layers can be components such as convolutional layers, LSTM layers and fully connected layers. The nodes also encapsulated the associated hyperparameters such as kernel size and activation function type. Notably, the algorithm used an indirect encoding. Their algorithm was benchmarked on an image dataset, chest x-rays \cite{rajpurkar2017chexnet}, and on the Wikipedia comment toxicity dataset. Although this algorithm evolved networks for text classification tasks, the networks were based on the LSTM paradigm and not a char-CNN approach. Further, their work was not applied on datasets commonly used to test char-CNNs. 
\section{PROPOSED ALGORITHM}
\subsection{Network Architecture Encoding}
Many evolutionary inspired network architecture search algorithms employ a direct encoding where the layers of the network are stated explicitly. This direct encoding approach often results in networks with questionable architecture arrangements, for example placing fully connected layers before convolutional layers, or worse, networks that are not fully formed or trainable. Further, direct encodings are susceptible to evolutionary operators being destructive on the network architecture. For example it is easy for a crossover operation to destroy the topology of a valid network architecture, resulting in wasted compute power. Special care needs to be taken when designing evolutionary operators for direct encodings. 
\par
Indirect encodings specify \emph{indirectly} how a network should be constructed through the use of program symbols, grammars or production rules. Networks can therefore be constructed in an incremental manner, ensuring that the structural integrity of a network is always maintained. Further, since evolutionary operations such as crossover are conducted on the data structure containing the program symbols to be executed, and not the network itself, the result will still generate a structurally valid neural network.
\par
An appropriate architecture encoding scheme is required to study the role of both depth and width (branching) in char-CNNs. The scheme must be complex enough to capture the properties of depth and width but also simple enough so as not to introduce additional variables of complexity. An encoding scheme representing a \emph{subset} of cellular encoding \cite{Gruau1994} operations is proposed. Cellular encoding draws inspiration from observation of cell divisions as seen in the biological domain. The encoding was originally designed to encode multi-layered perceptron (MLP) networks where the nodes in a MLP were represented as a biological cell to be operated on.
\par
\begin{figure}[h]
\centering
  \scalebox{0.10}{
        \includegraphics[width=1.0\textwidth]{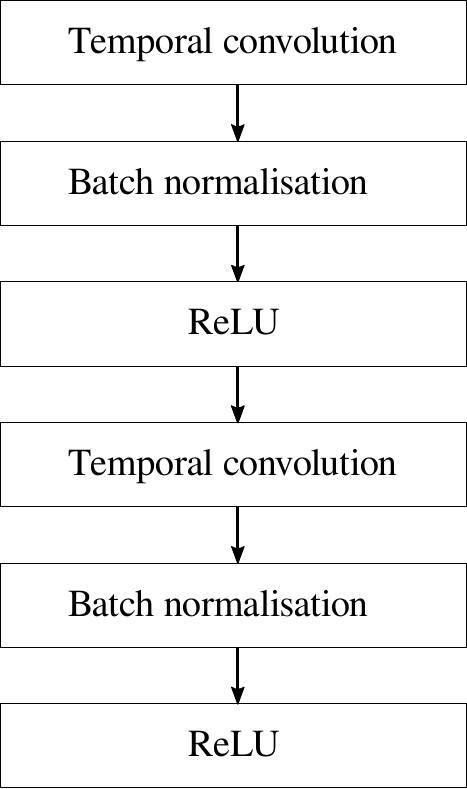}
    }\\
\caption{Network cell.}
\label{fig:cell}

\end{figure}
The chosen operations are the sequential (SEQ) and parallel (PAR) division operations. These two operations are a suitable choice as their application on a network's cell can construct network architectures of varying depth and width. A cell is defined as a convolutional block as used in \cite{Conneau2017} and presented in Figure \ref{fig:cell}. The SEQ and PAR operations are therefore applied to cells. Only one operation is applied to any given cell. 
\par
 An \emph{ancestor network} is defined as a cell coupled with an embedded input layer and an output layer. The output layer of the ancestor network consists of a temporal adaptive average pooling layer \cite{Lin} followed by a fully connected layer. The cross-entropy loss is propagated backwards through the ancestor network.
 \par
 In essence, the defined indirect encoding scheme represents a \emph{program} consisting of a sequence of operations to be performed on the cells of an ancestor network, making GP an appropriate and natural choice to evolve cellular encoded \emph{programs}.
\par
\subsection{Cellular Operations}
\subsubsection{SEQ operation} This operation produces a new cell (child cell) from the cell on which it operates (mother cell). The child cell is connected sequentially to the mother cell. The output of the mother cell is reassigned to the input of the child cell. The output of the child cell is, in turn, assigned to the original output of the mother cell.
\subsubsection{PAR operation} 
\begin{figure}[h]
\centering
\subfloat{
\scalebox{1.0}{
  \includegraphics[width=0.8\linewidth]{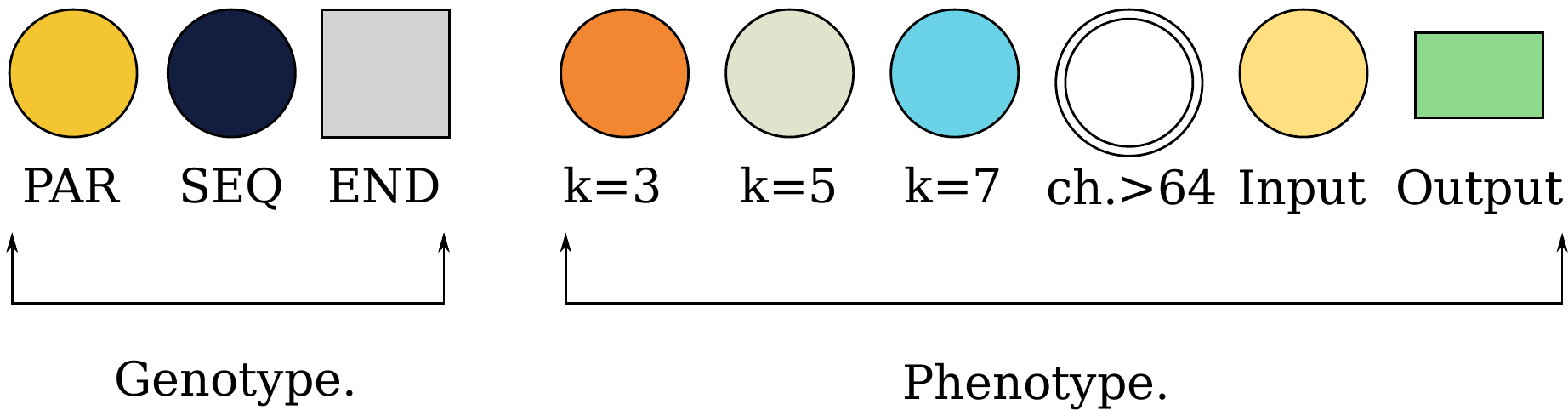}
  }
  }
\caption{Colour legend describing genotypes and phenotypes.}
\label{fig:color_legend}
\end{figure}

This operation also produces a child cell from the mother cell. However, the child cell is assigned a new kernel size and connected in parallel to the mother cell. The kernel size is selected from a list of possible values. The chosen list includes kernel sizes of 3, 5 or 7. These values are optimum with regards to char-CNNs \cite{Conneau2017}. The selection is based on a shift-right method. If the mother cell has a value of 3, then the child kernel size is assigned a value of 5. A mother cell with a kernel size of 7 will result in a child cell with a kernel size of 3. This method is deterministic and required in order to ensure that the same phenotype can be constructed consistently from a given genotype. The input and output destination of the child cell are assigned the same input and output destination cells as the mother cell. This implies that if the mother and child cell are connected to another cell and not the output layer, then a concatenation operation is to be performed in the destination cell’s input. The concatenation operation is simply the stacking of each incoming cell’s channels on top of each other. For example, if two cells, each having 64 output channels, connect to the same destination cell, then the destination cell will have 128 input channels. In order to make the concatenation of input channels possible, due to the varying temporal lengths resulting from different kernel sizes, padding with a zero value is used to extend the temporal dimension to equal size lengths. A legend is provided in figure \ref{fig:color_legend} to aid in the description of genotypes and phenotypes for the remainder of this work.

\vspace{-8mm}
\begin{figure}[H]
\centering
\subfloat[Ancestor.]{
\scalebox{0.36}{
\hspace{12mm}
  \includegraphics[width=0.20\linewidth]{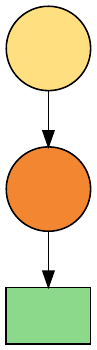}
  }
  \label{fig:ancestor_phenotype}
  }
\hspace{5mm}
\subfloat[SEQ operation.]{
  \includegraphics[width=0.25\linewidth]{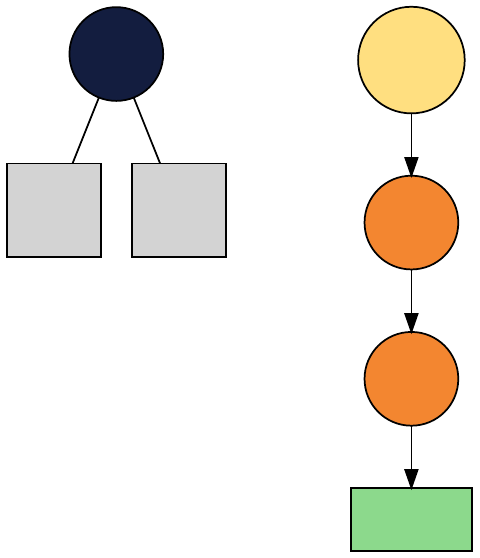}
  \label{fig:seq_smallest}
  }
\hspace{3mm}
\subfloat[PAR operation.]{
  \includegraphics[width=0.30\linewidth]{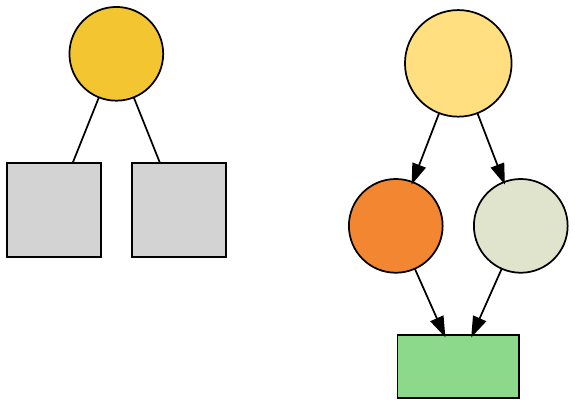}
  \label{fig:par_smallest}
  }
\label{I_V}
\caption{Smallest possible phenotypes.}
\label{fig:smallest_networks}
\end{figure}
When discussing the genotype, each cellular operation is represented by a colour-coded circle, as indicated in figure \ref{fig:color_legend}. The cells in a network (phenotype) are represented by coloured circles where the colour is related to the assigned kernel size. A double circle represents a cell with more than the default 64 input channels, indicating that a concatenation operation has occurred previously along the path of the input layer to the current cell. An example of the smallest phenotypes that can be constructed from the ancestor phenotype is displayed in figure \ref{fig:smallest_networks}. Each phenotype is displayed alongside its relevant genotype. It can be seen that a SEQ operation contributes to the depth of a network and a PAR operation contributes to the width of a network.

\subsection{Surrogate Models}
To aid in reducing the computational time for evaluating char-CNNs, this work makes use of half precision (16-bit) training. This work was conducted over four retail RTX 2070
cards. These RTX cards contain tensor cores\footnote{https://www.nvidia.com/en-us/data-center/tensorcore/} that are significantly faster than CUDA cores. Tensor cores are only activated under certain conditions, one of which is using half precision training. Nvidia states that tensor cores reduce the training time of deep neural networks by a factor between 2 and 4 depending on the task. There is a slight trade-off in reduced accuracy. We refer to models trained using half precision as surrogate models. The loss in accuracy performance is not relevant when using evolutionary deep learning techniques as we are only interested in evolving surrogate phenotypes and then using full resolution (32-bit) training for the fittest phenotype. An added benefit of using lower resolution training is that the GPU’s memory is effectively doubled. However, this poses the problem of producing surrogate phenotypes that fill the entire available GPU memory and implies that the full resolution version of the phenotype will be too large to fit in the GPU’s available memory. To overcome this potential problem, the high resolution phenotype is always trained over two GPU’s. Nvidia's Apex AMP library\footnote{https://github.com/NVIDIA/apex} was used to enable half precision training.

\subsection{Algorithm Overview}
The proposed algorithm evolves genotypes, represented as GP trees containing program symbols, by using evolutionary crossover and mutation operators. These program symbols represent actions that are to be performed when constructing the network architecture. The evolved genotypes are decoded, by executing the program symbols, to construct phenotypes which represent trainable network architectures. The phenotypes are trained using the backpropogation algorithm, and their final validation accuracy is used to evaluate the fitness of the phenotype relative to all other phenotypes in the population. The use of surrogate models enables the phenotypes to be trained significantly faster. At the end of the evolutionary process, the fittest surrogate phenotype is automatically located and trained as a non-surrogate phenotype. The trained non-surrogate phenotype is then evaluated on the test set. The algorithm terminates by presenting the genealogy of the fittest phenotype for analysis.
\begin{algorithm}[h]
    \caption{Proposed algorithm.}
    \textbf{begin}\;
    \quad $seed \leftarrow$ Assign\ next\ seed\ from\ list.
    \quad $population \leftarrow$ genotypes\ with\ depth\ range [1,3].
    \While{$not\ maximum\ generations$}{%
        \ForEach{$genotype \in population$}{%
          GPU $\leftarrow phenotype \leftarrow decode(genotype)$\;
           \While{$phenotype\ not\ accepted\ by\ GPU$}{%
                $genotype \leftarrow genotype\ depth\ halved$\;
                GPU $\leftarrow phenotype \leftarrow decode(genotype)$\;
           }
          $evaluate(genotype,\ reduced\ train.\ set,\ val.\ set)$\;
        }
        $elite \leftarrow$ fittest\ from\ population\;
        $selected \leftarrow tournament(population)$\;
        $offspring\ population \leftarrow crossover(selected)$\;
        $population \leftarrow mutate(offspring\ population)$\;
        $limit(population \cup elite)$;
     }
     $fittest \leftarrow$ population\;
     GPU $\leftarrow 32$-$bit\ phenotype \leftarrow decode(fittest)$\;
     $evaluate(fittest,\ full\ train.\ set,\ test\ set)$\;
    \textbf{end}\;
    \label{alg:evolutionary_process}
\end{algorithm}

The framework of the proposed algorithms is presented in Algorithm 1. The algorithm is assigned a randomly generated seed. The population is then initialised as detailed in section \ref{sec:pop_init}. During the evolutionary process, each genotype in the population is first decoded into its half resolution (16-bit) phenotype, which represents a trainable CNN. The phenotype is uploaded to the GPU. If the phenotype is too large to fit in the memory the GPU, the phenotype is destroyed and the genotype is reduced in depth by a factor of two. The modified genotype is then decoded to a new phenotype and again transferred to the GPU. This process is repeat until the a phenotype is accepted. The phenotype on the GPU is then trained using the reduced training set and evaluated on the validation set. After each genotype and its corresponding phenotype has been evaluated, the elite population is constructed from the fittest 10\% of the population. These elite individuals are copied over to the next generation without any modification to them. A tournament selection function is then applied to the entire population. The selected individuals are used to produced new offspring via the single point crossover operation. A uniform mutation operation is then applied to the offspring. A new  population is then constructed consisting of the newly generated offspring and the elite population. This process is repeated till the maximum number of generations has been achieved. The fittest individual from the final population is selected and decoded as a full resolution (32-bit). This full resolution phenotype is then trained using the full training set and evaluated on the test set.

\subsubsection{Evolutionary operators}
The proposed method uses single point crossover operations. Two selected genotypes are crossed over at a randomly selected position, resulting in two offspring genotypes. The mutation operation selects a random position in the genotype and connects a randomly generated sub-tree. This produces a small change in the genotype program which translates to a physical alteration in the construction of the phenotype. The crossover and mutation operations described were selected for their simplicity as future research will investigate  novel  crossover and mutation methods.
\subsubsection{Population Initialisation}\label{sec:pop_init}
\section{Experimental Design}
\subsection{Peer Competitors}
There is no research using EDL to evolve char-CNNs in the literature, therefore to test the efficacy of the SurDG-EC algorithm, a comparison is conducted against an algorithm using the same encoding but with no evolutionary operators (SurDG-Random). The fittest evolved phenotype located by both the SurDG-EC algorithm and SurDG-Random are retrained as higher resolution phenotypes and compared against three hand-crafted state-of-the-art char-CNNs: Zhang et al’s \cite{Zhang2015} small and large models (Small Char-CNN, Large Char-CNN) and Conneau et al’s \cite{Conneau2017} model (VDCNN-29). All these models are pure character-level models, meaning that there is no data augmentation or pre-processing of the input text data. A comparison is also made against three word-level CNNs using the popular Word2vec model. The peer competitor results are reported from \cite{Zhang2015} and \cite{Conneau2017}.
\subsection{Benchmark Datasets}
Zhang et al.’s \cite{Zhang2015} work on the first char-CNN was tested on eight datasets as listed in Table \ref{tab:datasets_evolution}. The datasets are considered to be the standard for evaluating text classification performance of char-CNNs.
\begin{table}[H]
\centering
\scalebox{0.95}{
  \begin{tabular}{lcrrr}
  \hline
    Dataset                 & Classes & Train & Validation & Test\\
    \hline
    AG’s News               & 4     & 112,852 & 7,148 & 7,600 \\
    Sogou News              & 5     & 397,058 & 52,942 & 60,000 \\
    DBPedia                 & 14    & 497,777 & 62,223 & 70,000 \\
    Yelp Review Polarity    & 2     & 524,414 & 35,586 & 38,000 \\
    Yelp Review Full        & 5     & 603,571   & 46,429 & 50,000 \\
    Yahoo! Answers          & 10    &  1,342,465 & 57,535 & 60,000 \\
    Amazon Review Full      & 5     &  2,465,753 & 534,247 & 650,000 \\
    Amazon Review Polarity  & 2     &  3,240,000 & 360,000 & 400,000 \\
    \hline
  \end{tabular}
  }
  \caption{Datasets by training, validation and test splits.}
  \label{tab:datasets_evolution}
\end{table}
The AG’s News dataset is regarded as a challenging dataset to classify because it contains a small number of instances. This dataset was chosen for this work as there is still potential in improving the classification accuracy over it. The remainder of the datasets were not used in the evolutionary process; however, they were used to evaluate the ability of the fittest evolved phenotype to generalise over the other unseen datasets. It is noted that neither of the original eight datasets had a validation set. Therefore the original training sets were each split into a reduced training set and a validation set. The split ratio was kept the same as between each original training set and test set. The original test sets remained unaltered.

An analysis of the instances in each dataset is listed in Table \ref{tab:datasets_stats}. Zhang et al.’s \cite{Zhang2015} original char-CNN used a temporal length of 1014 characters. Most of the instances in the AG’s News dataset are closer to 256 characters. Setting the temporal length to 1014 would imply unnecessary padding and convolutional operations, resulting in wasted compute power and time. Therefore this work used a maximum sentence length of 256 characters. This aided in improving model training times without the loss of significant discriminative information from each instance with the regards to the AG’s News dataset. It is noted that the other seven datasets have a mean length greater than 256, implying that important sentence data may have been truncated when the evolved architecture was evaluated on them.

\begin{table}[h]
\centering
  \scalebox{0.95}{
  \begin{tabular}{lrrr}
    \hline
    Dataset & Mean & Minimum & Maximum \\
    \hline
    AG's News               & 236$\pm$66 & 100 & 1,012 \\
    Sogou News              & 2,793$\pm$3,338 & 40 & 185,674 \\
    DBPedia                 & 301 $\pm$139 & 12 & 13,574 \\
    Yelp Review Polarity    & 725 $\pm$669 & 10 & 8,787 \\
    Yelp Review Full        & 732$\pm$664 & 10 & 5,849\\
    Yahoo! Answers          & 520$\pm$577 & 12 & 8,191 \\
    Amazon Review Full      & 439$\pm$240 & 96 & 1,884 \\
    Amazon Review Polarity  & 430$\pm$237 & 71 & 1,981 \\
    \hline
  \end{tabular}
 }
  \caption{Sentence lengths.}
  \label{tab:datasets_stats}
\end{table}
\begin{figure*}[t]
\vspace{-9mm}
\centering
\scalebox{0.85}{
        \includegraphics[width=1.0\textwidth]{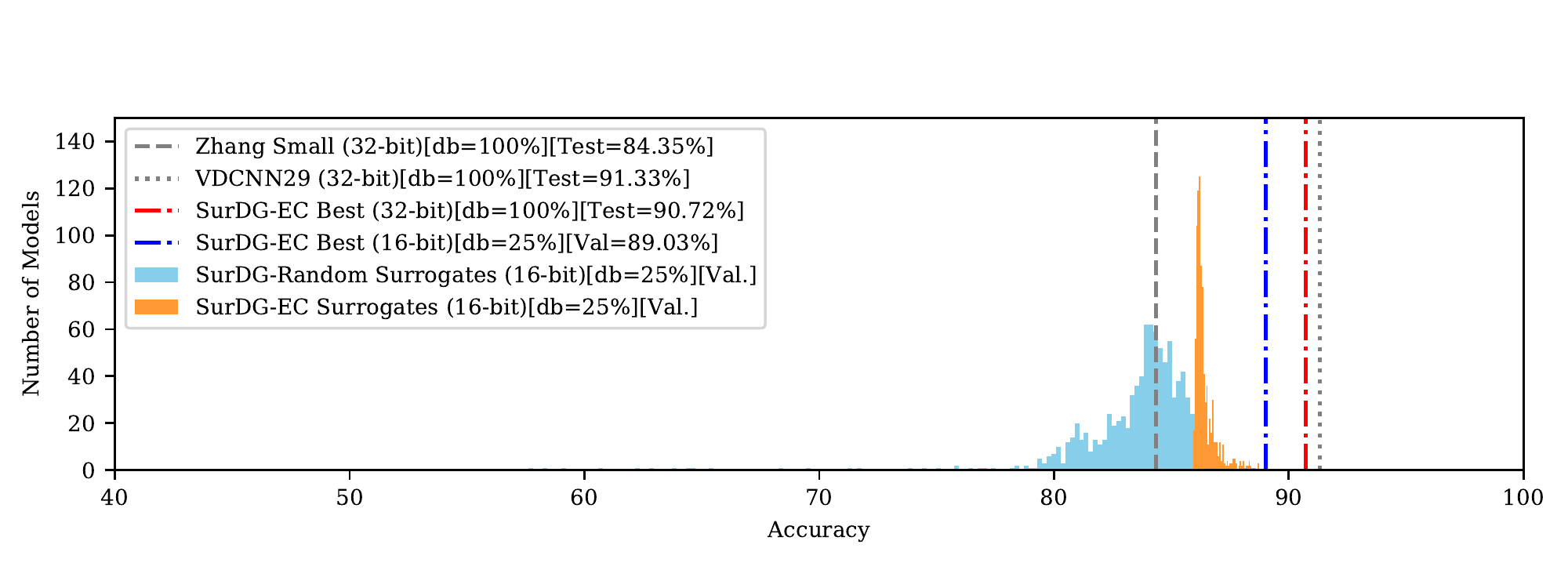}
    }
    \vspace{-5mm}
    \caption{AG's News: Distributions of validation accuracies including competitor test accuracies.}
    \label{fig:accuracy_histogram}
\end{figure*}
\vspace{-5mm}
 \subsection{Parameter Settings}
The parameters for the components of the experiment are listed in Table \ref{tab:parameters}. Thirty runs were conducted for each algorithm, where each run was assigned a single unique seed. Every surrogate phenotype was trained for ten epochs. This approach ensured that evolutionary pressure to converge quickly to high accuracy was applied to all evolved phenotypes. Limiting the epochs to 10 also aided in reducing the run time of the experiment. The batch size, initial learning rate and learning schedule were assigned values as in \cite{Zhang2015} and \cite{Conneau2017}. These values are considered best practice. An AMP profile of O2 was used to generate the low-resolution surrogate models. The selected optimiser was a stochastic gradient descent (SGD) function. SGD was used in both \cite{Zhang2015} and \cite{Conneau2017}. The initial settings of the cellular cells were the same as the convolutional blocked used in \cite{Conneau2017}.
\par
Each run consisted of 30 generations over a population size of 30. There is much research on determining what is the best ratio of generation size to population size for evolutionary algorithms, however not with regards to the domain of evolutionary deep learning, where computation times are significantly longer than most other domains. A pragmatic approach was taken by setting the number of generations and population size to be equal to each other with the assumption that any effect of increasing one over the other is neutralised. The elitism, crossover and mutation rate settings were based on common values found in the literature \cite{10.5555/138936}. The mutation growth depth was set to a maximum of size two, ensuring that a mutation event did not significantly change the structure of the genotype to prevent the possibility of destroying the quality of the phenotype. The maximum tree depth was set to 17, as recommended in \cite{10.5555/138936}. The GPU rejection re-initialisation was set to a maximum of depth 4. A low value was chosen to increase the odds of the GPU accepting the model on any further submission attempts. The fitness function was the same as in \cite{Zhang2015} and \cite{Conneau2017}. The overall objective of the evolutionary process was to maximise the validation accuracy.
\vspace{3mm}
\begin{table}[h]
  \centering
  \scalebox{0.95}{
  \begin{tabular}{lr}
   \hline
    Parameter                 & Value\\
    \hline
    Run count              & 30 \\
    Random seeds    & Unique per Run \\
    \emph{Deep Learning:}     &  \\
    Epochs                 & 10 \\
    Batch Size    & 128 \cite{Zhang2015},\cite{Conneau2017} \\
    Initial Learning Rate    & 0.01 \cite{Conneau2017} \\
    Momentum & 0.9 \cite{Conneau2017}\\
    Learning Schedule & Halve every 3 epochs \cite{Zhang2015} \\   
    Weight Initialisation & Kaiming\cite{Kaiming2018a}\cite{Zhang2015},\cite{Conneau2017}\\
    \emph{Surrogates:}       &  \\
    AMP Profile \\ 
    AMP Optimiser & SGD \cite{Zhang2015} \\
    Training Data Usage & 0.25\\
    \emph{Character-level Model:}       &  \\
    Alphabet & Same as in \cite{Zhang2015} and \cite{Conneau2017} \\   
    Max Sentence Length & 256\\
    \emph{Cellular Cell - Initial}       &  \\
    In and Out Channels  & 64 \cite{Conneau2017} \\ 
    Activation Function & ReLu \cite{Conneau2017}  \\
    Kernel size & 3 \cite{Conneau2017} \\
    Stride & 2 \cite{Conneau2017} \\
    Padding & 1 \cite{Conneau2017} \\
    \emph{Evolutionary:}   &  \\
    Number of Generations & 30  \\
    Population Size & 30\\
    Elitism & 0.1 \\
    Crossover Probability & 0.5 \\
    Crossover Type & Single Point \\
    Mutation Probability & 0.1 \\
    Mutation Distribution & Uniform \\
    Mutation Growth Type & Grow \\
    Mutation Tree Growth Size& [1,2] \\
    Tournament Selection Size & 3 \\
    Primitives & \{SEQ,\space PAR\} \cite{Gruau1994} \\
    Terminals & \{END\} \cite{Gruau1994} \\
    Max Tree Depth & 17 \cite{10.5555/138936} \\
    Initial Tree Depth & [1,3] \\
    Initial Tree Growth & Half and Half \\
    GPU Rejection Tree Re-initialisation & [1,Half previous] \\
    Fitness Function & max(validation accuracy) \cite{Zhang2015} \cite{Conneau2017}\\
    \\
    \hline
  \end{tabular}
  }
  \caption{Parameter settings.}
  \label{tab:parameters}
\end{table}Momentum
\vspace{-7mm}
\subsection{Statistical Tests}
Thirty random seeds were generated before any algorithms or searches were conducted. Each seed was used to conducted one run of the SurDG-EC algorithm \emph{and} one run of SurDG-Random separately, in different application processes. It is noted that both runs were conducted on the same hardware. Both the SurDG-EC algorithm and SurDG-Random were conducted on the exact same reduced training set. This implies a paired coupling of observations. As the distribution of the samples is not know and the observations are paired, a Wilcoxon signed-rank test was used.
\par
The fittest phenotype from each run of SurDG-Random was compared against the fittest phenotype from each SurDG-EC run. This translates to the 30 fittest phenotypes located by SurDG-Random compared to the 30 fittest phenotypes located by the SurDG-EC algorithm. A significance level of 0.05 was applied. The null hypothesis was that the distribution of the observations of both methods came from the same population.
\section{Results and analysis}
\subsection{Overall results}

The aggregated validation accuracies of the surrogate phenotypes generated by both SurDG-Random and the SurDG-EC algorithm are presented in Figure \ref{fig:accuracy_histogram}. The distribution of the validation accuracies sampled by SurDG-Random, presented in light blue, represents a normal distribution. This result indicates that sufficient samples were extracted to represent the overall population landscape. It is noted that SurDG-Random located a negligible number of surrogate phenotypes with validation accuracies of less than 60 percent. These were omitted to improve visual clarity in the figure. Zhang et al.’s \cite{Zhang2015} original char-CNN model has a \emph{test} accuracy close to the mean of the surrogate phenotypes validation accuracy located by SurDG-Random. This result suggests, with caution, that the search space may contain surrogate phenotypes that are on average, similarly performant as Zhang et al.’s \cite{Zhang2015} original char-CNN model.
\par
The distribution presented in orange represents the validation accuracies sampled by the SurDG-EC algorithm. The distribution consists of the final population of each run of the SurDG-EC algorithm, representing 900 surrogate phenotypes. It is noted that the distribution loosely represents half a normal distribution. This observation is expected as the lower bound is clipped due to only the fittest surrogate models surviving up to the end of an evolutionary run. It is easily observable that the mean of this distribution is shifted from the mean of SurDG-Random’s distribution. It can be seen that the right-hand tail of the SurDG-EC distribution extends further than the right-hand tail of SurDG-Random’s distribution. The SurDG-EC algorithm has located higher accuracy models compared to those found by SurDG-Random. Application of the  Wilcoxon signed-rank test resulted in a rejected null hypothesis implying that the distributions were significantly different.
The fittest surrogate phenotypes located by the SurDG-Random and SurDG-EC algorithm achieved validation accuracies of 87.57\% and 89.03\% respectively. The genotype that generated the fittest SurDG-EC surrogate phenotype was used to construct a higher resolution phenotype. This higher resolution phenotype was then trained on 100\% of the reduced training set and evaluated on the same test set used in \cite{Zhang2015} and \cite{Conneau2017}. It is noted that both Zhang et al.’s and Conneau et al.’s models were trained on the original training set of 120,000 instances whereas the SurDG-EC algorithm could only be trained on the reduced training set of 112,852 instances in order not to introduce training bias. This decision potentially limited the accuracy during the training of the SurDG-EC algorithm and gave an unfair advantage to Zhang et al.’s and Conneau et al.’s models. Regardless, the final test accuracy achieved by the trained full resolution phenotype was 90.72\% as indicated by the red dashed line in figure \ref{fig:accuracy_histogram}. The full resolution phenotype outperformed Zhang et al.’s model by 6.37\% and under-performed Conneau et al.’s model by only 0.61\%.
\begin{table}[H]
  \centering
  \scalebox{0.9}{
  \begin{tabular}{lrr}
  \hline
    Measure & SurDG-Random & SurDG-EC  \\
    \hline
    Validation Accuracy (Best) & 83.26$\pm$3.8\space(\textbf{87.57})& 86.42$\pm$0.42(\textbf{89.03}) \\
    Train Time (Seconds) & 141.60$\pm$132.42& 134.11$\pm$90.59 \\
    Parameter Count & 1,034k$\pm$501k & 441k$\pm$77k \\
    SEQ operations & 3,716 & 3,046 \\
    PAR operations & 3,595 & 634 \\
    Crossover operations (30 runs) & - & 5,732 \\
    Mutation operations (30 runs) & - & 2,358 \\
    \hline
  \end{tabular}
  }
  \caption{AG's News: SurDG-Random vs SurDG-EC.}
  \label{tab:ag_news_stats}
\end{table}
\subsection{Comparison of SurDG-random and SurDG-EC}
The average validation accuracies achieved by both the SurDG-Random and SurDG-EC algorithm are listed in Table \ref{tab:ag_news_stats}. Average accuracies of 83.26\% for SurDG-Random and 86.42\% for the SurDG-EC algorithm were attained. The average training time of surrogate phenotypes from both methods are roughly similar at 141 and 134 seconds. SurDG-Random has a higher standard deviation, indicating a wider spread of training times from the mean train time when compared to the SurDG-EC algorithm. This finding is not surprising as SurDG-Random is likely to have covered a wider search area, indicating a broader range of trainable parameter sizes. The SurDG-EC algorithm found better solutions, in general, in less time than SurDG-Random.
\par
The ratio between SEQ and PAR operations executed during SurDG-Random was approximately 50:50 at 3,716 SEQ and 3,595 PAR executions. This ratio is expected as each operation has a 50\% chance of being selected when constructing the genotype. Interestingly, the SurDG-EC algorithm has a higher number of SEQ operations to PAR operations executed. This observation indicates that SEQ operations played a more prominent role in achieving higher validation accuracies during the evolutionary process. In general, more SEQ operations hint at deeper networks, agreeing with the findings in \cite{Le}, that deeper character-level CNNs are more accurate than shallow ones.
\par
The number of crossover and mutation operations executed were 5,732 and 2,358, respectively, for the SurDG-EC algorithm as listed in Table \ref{tab:ag_news_stats}. Running the SurDG-EC algorithm 30 times with a population of size 30 over 30 generations gives a potential of 27,000 model evaluations that could be performed. With an elite population of 10\%, the number reduces to approximately 24,390 as any model in the elite population is only evaluated once. The crossover operator is applied pairwise to each individual in the population and its neighbour. This technique limits the maximum number of crossover operations to 12,195, assuming a crossover rate of 100\%. However, the crossover probability is set at 50\%, limiting the maximum number of crossover operations to 6,097, which is close to the reported value of 5,732 operations. The mutation rate of 10\% resulted in 2,358 mutation operations being executed. This value translates to 10\% of the possible 24,390 potential evaluation operations. The reported crossover and mutation values are consistent with their settings. This observation highlights one aspect of the veracity of the SurDG-EC algorithm, namely that the correct percentage of evolutionary operations have been performed.
\vspace{-5mm}
\begin{figure}[H]
  \centering
\subfloat[SurDG-Random.]{
  \hspace{-7mm}
\scalebox{0.63}{
  \includegraphics[width=0.9\linewidth]{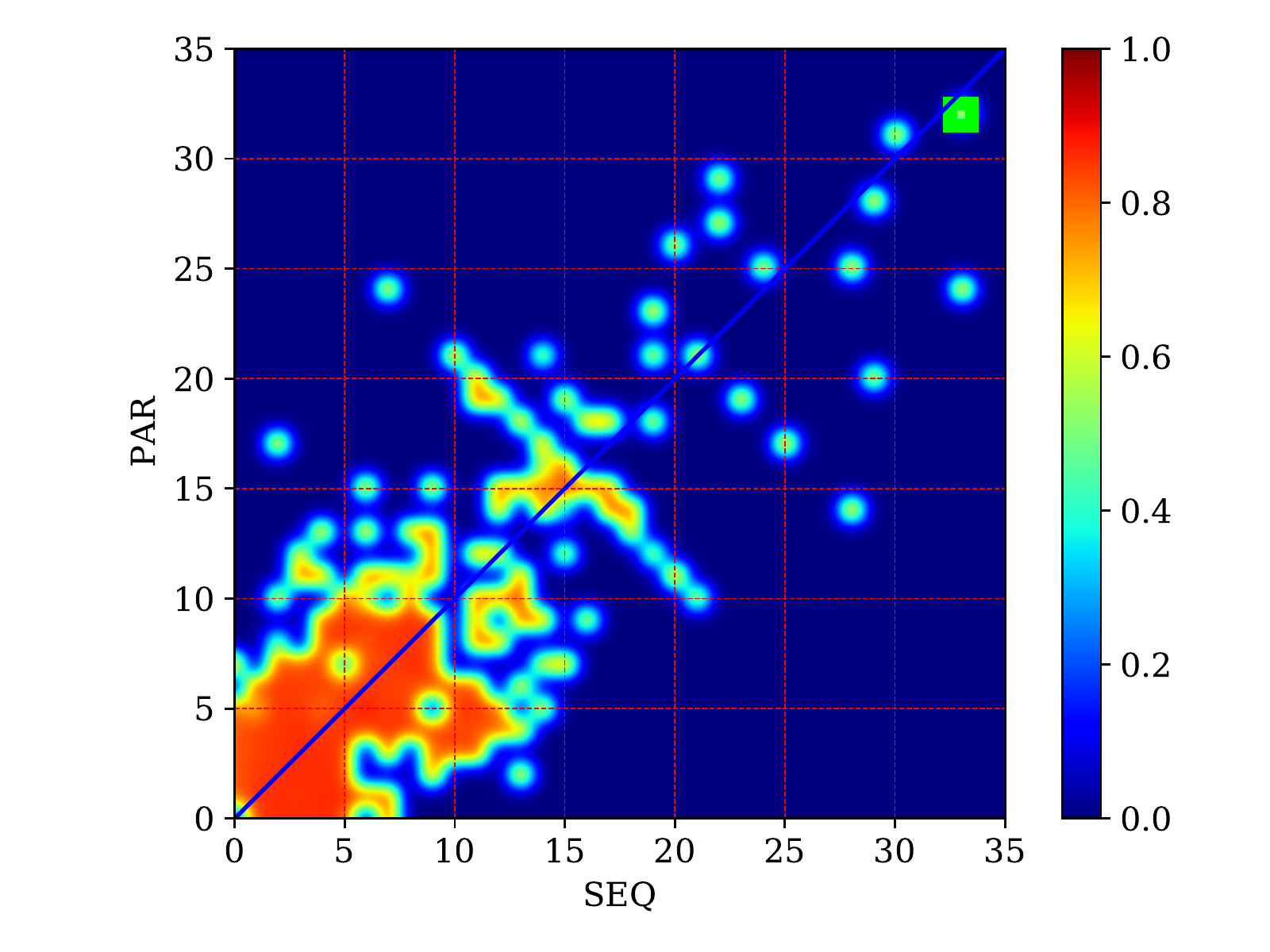}
 \label{fig:heatmap_a}
  }}
\subfloat[SurDG-EC.]{
\hspace{-8mm}
\scalebox{0.63}{
  \includegraphics[width=0.9\linewidth]{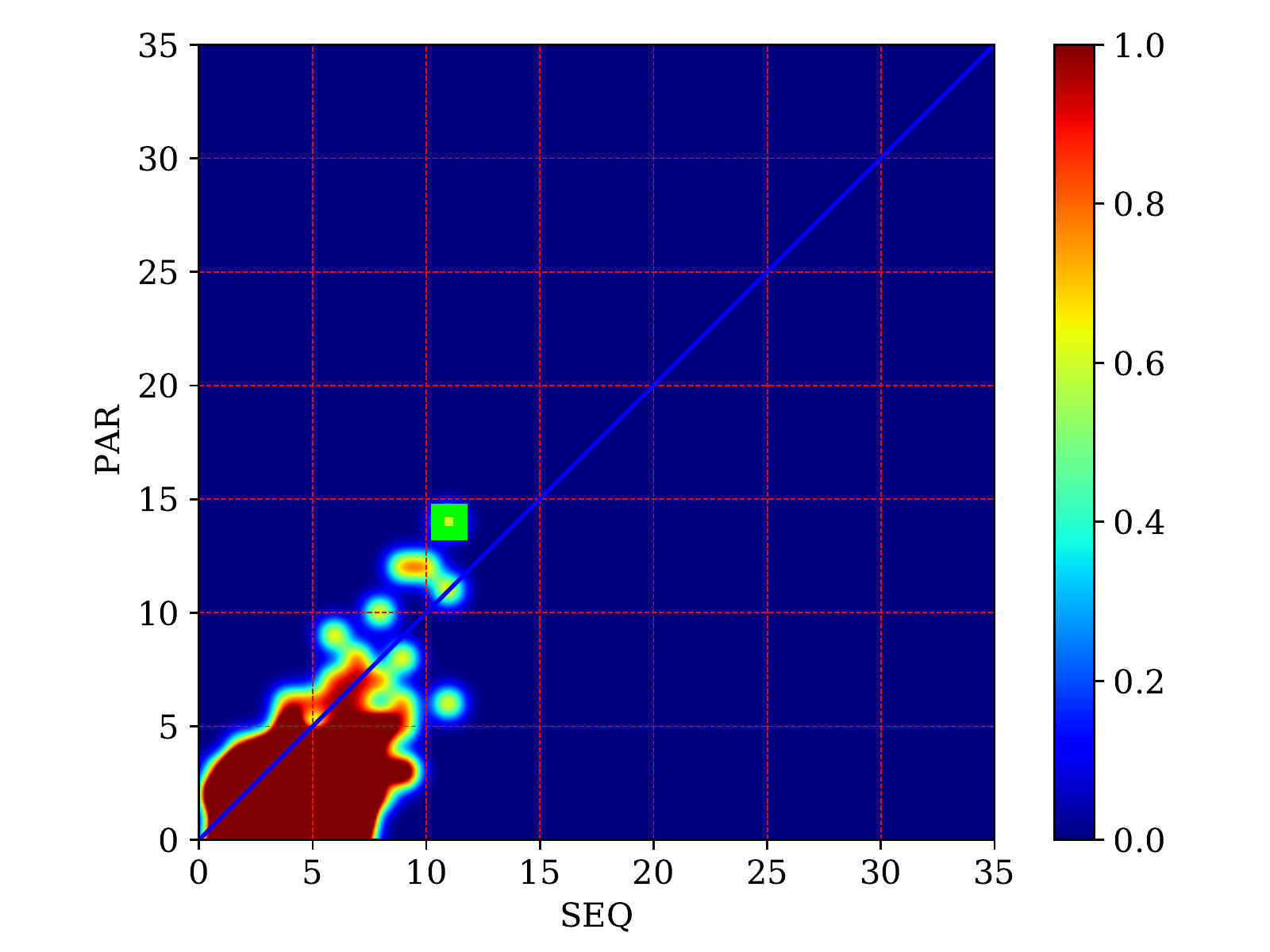}
   \label{fig:heatmap_b}
    }
  }
\caption{Density of number of SEQ vs PAR operations.}
\label{fig:density_heatmap}
\end{figure}
\vspace{-3mm}
The distribution of the number of SEQ and PAR operations that constitutes each phenotype is presented in Figure \ref{fig:density_heatmap} with sub-figure \ref{fig:heatmap_a} representing SurDG-Random and sub-figure \ref{fig:heatmap_b} the SurDG-EC algorithm. It can be seen that all the surrogate phenotypes located by SurDG-Random, cluster around the blue diagonal centre line. This behaviour is due to each cellular operation (SEQ and PAR) having a 50:50 chance of being selected when constructing the genotype. It is doubtful that a phenotype with 20 SEQ operations and 1 PAR operation would be located without the aid of an external force such as evolutionary pressure. It can be observed that the number of surrogate models located by SurDG-Random becomes sparse with the increase of SEQ and PAR operations. This is due to the models getting larger and not being able to be loaded into the GPU's memory. An analysis of the diagonal heat cluster located near the centre of the image confirms the existence of models that were initially rejected, modified and then re-uploaded to the GPU. This also explains the high concentration of phenotypes with SEQ and PAR operations between 0 and 10 operations as, again, any GPU rejected model is restricted to smaller tree depths and reloaded to the GPU.
\par
The fittest phenotype located by SurDG-Random, is highlighted with a lime green square and located in the first quadrant in Figure \ref{fig:heatmap_a}. It is interesting that this model has a large number of SEQ and PAR operations and thereby possibly a large number of parameters. However, the number of parameters is not only related to the number of SEQ and PAR operations but also the order in which those operations are executed. For example, a network constructed of 10 SEQ operation executed and then 1 PAR operation executed, will have less trainable parameters than a network constructed from 1 PAR operation executed and then 10 SEQ operations executed. This is due to the concatenation of channels from the PAR operation which will increase the number of channels, and the subsequent SEQ operations will propagate those increased channel numbers down the network stack, increasing the number of trainable parameters. It is noted that this located surrogate phenotype has roughly 18 million parameters. The SurDG-EC algorithm located a high concentration of phenotypes consisting of SEQ operations numbering between 0 and 10, and PAR operations numbering between 0 and 5. This finding indicates that SEQ operations played a dominant role during the evolutionary process. The fittest model is highlighted in lime green and located in the third quadrant near the blue centre line of Figure \ref{fig:heatmap_b}. It is of interest that both the \emph{fittest} phenotypes found by SurDG-Random and SurDG-EC algorithm are located around the centre line. 32 SEQ and 33 PAR operations were executed to produce the phenotype found by SurDG-Random. 14 SEQ and 11 PAR operations were executed to produce the phenotype found by the SurDG-EC algorithm. This implies that each phenotype had an almost equal ratio of PAR and SEQ operations applied to it. This may be an indication that \emph{both} PAR and SEQ operations are important, alluding to the conclusion that width and depth may potentially be an important combination for char-CNNs.
\par
\subsubsection{SurDG-Random}
\begin{figure}[h]
  \centering
  \scalebox{0.4}{
        \includegraphics[width=1.0\textwidth]{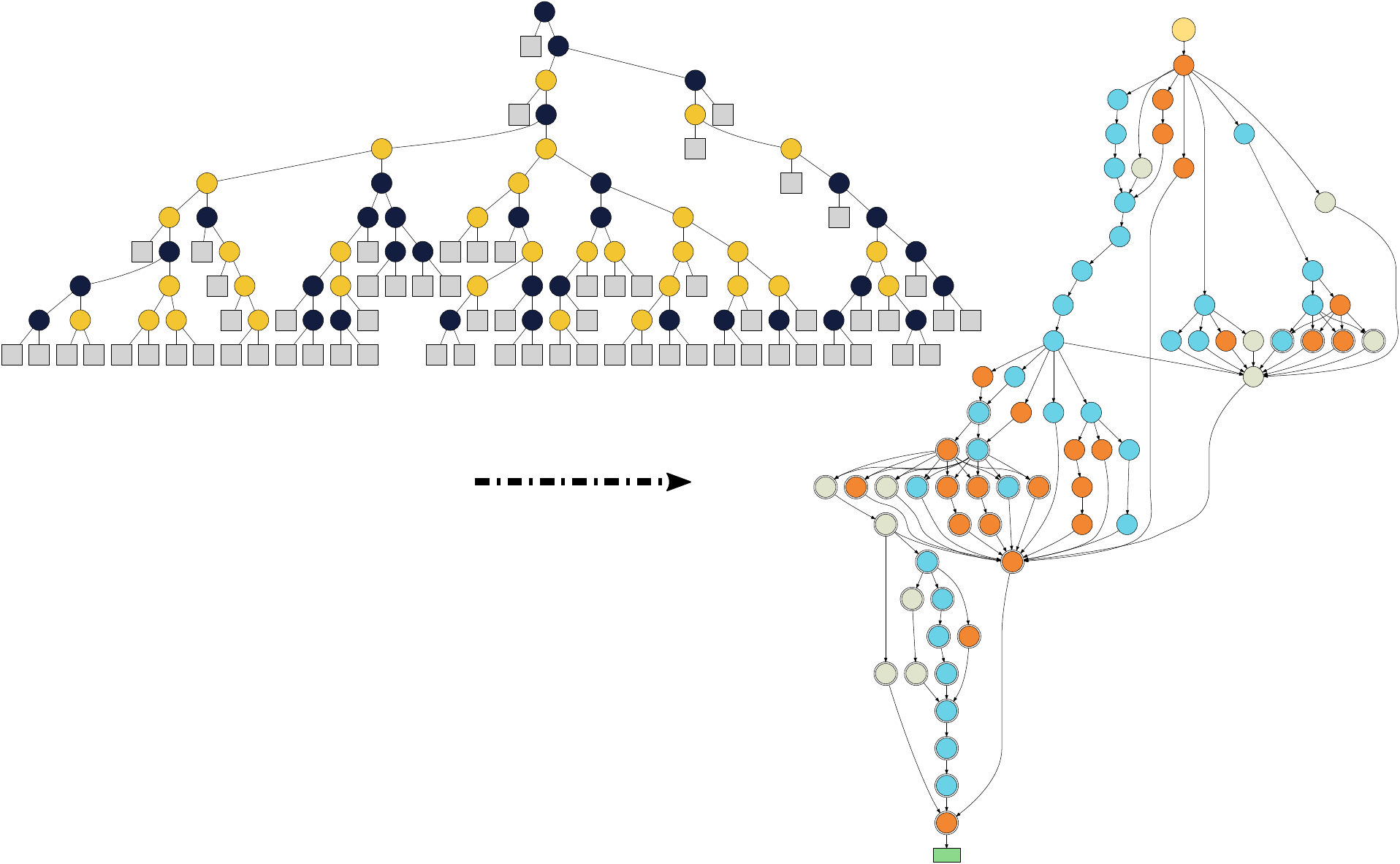}
    }
    \caption{Fittest genotype/phenotype found by SurDG-Random.}
    \label{fig:fittest_random_genotype_phenotype}
\end{figure}

The fittest genotype and corresponding phenotype found by SurDG-Random is presented in Figure \ref{fig:fittest_random_genotype_phenotype}. The phenotype has an almost diagonally-mirrored symmetry to it.
\par
There are a few interesting properties to note about the genotype and phenotype. Firstly the genotype has little visual representation to the phenotype, implying that it is difficult to predict the effect that a change in the genotype may have on the phenotype. This could potentially be a limitation of the chosen encoding scheme. If a small change in the genotype results in a significant structural change in the phenotype, the fitness landscape may not be smooth enough for evolutionary computation techniques to perform any better than a random search. There is another interesting property to note about this phenotype. There are a few critical nodes present almost as if the phenotype consists of a collection of smaller phenotypes connected by these critical nodes. It is noted that the bottom part of the network has a wide segment, followed by a narrow and deep segment of the network. This same property is present in the fittest phenotype located by the SurDG-EC algorithm, that will be discussed further on.
\subsection{Analysis of SurDG-EC}
\begin{figure}[h]
  \scalebox{0.5}{
\centering
\includegraphics[width=1.0\textwidth]{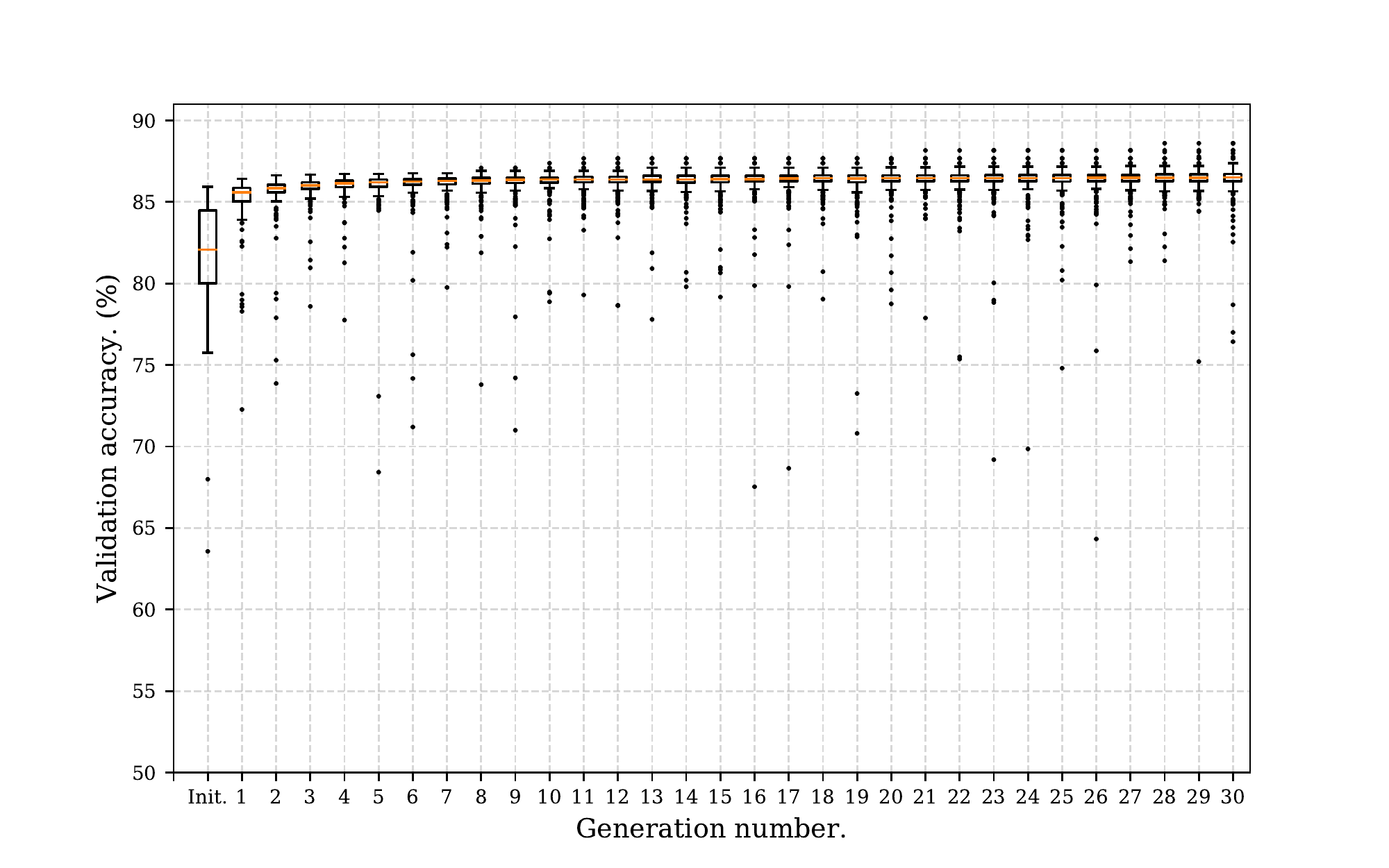}
        }
    \caption{AG's News: SurDG-EC performance over generations.}
    \label{fig:generation_box_plot_all_runs}
\end{figure}

\begin{figure}[H]
  \centering
  \scalebox{0.2}{
        \includegraphics[width=1.0\textwidth]{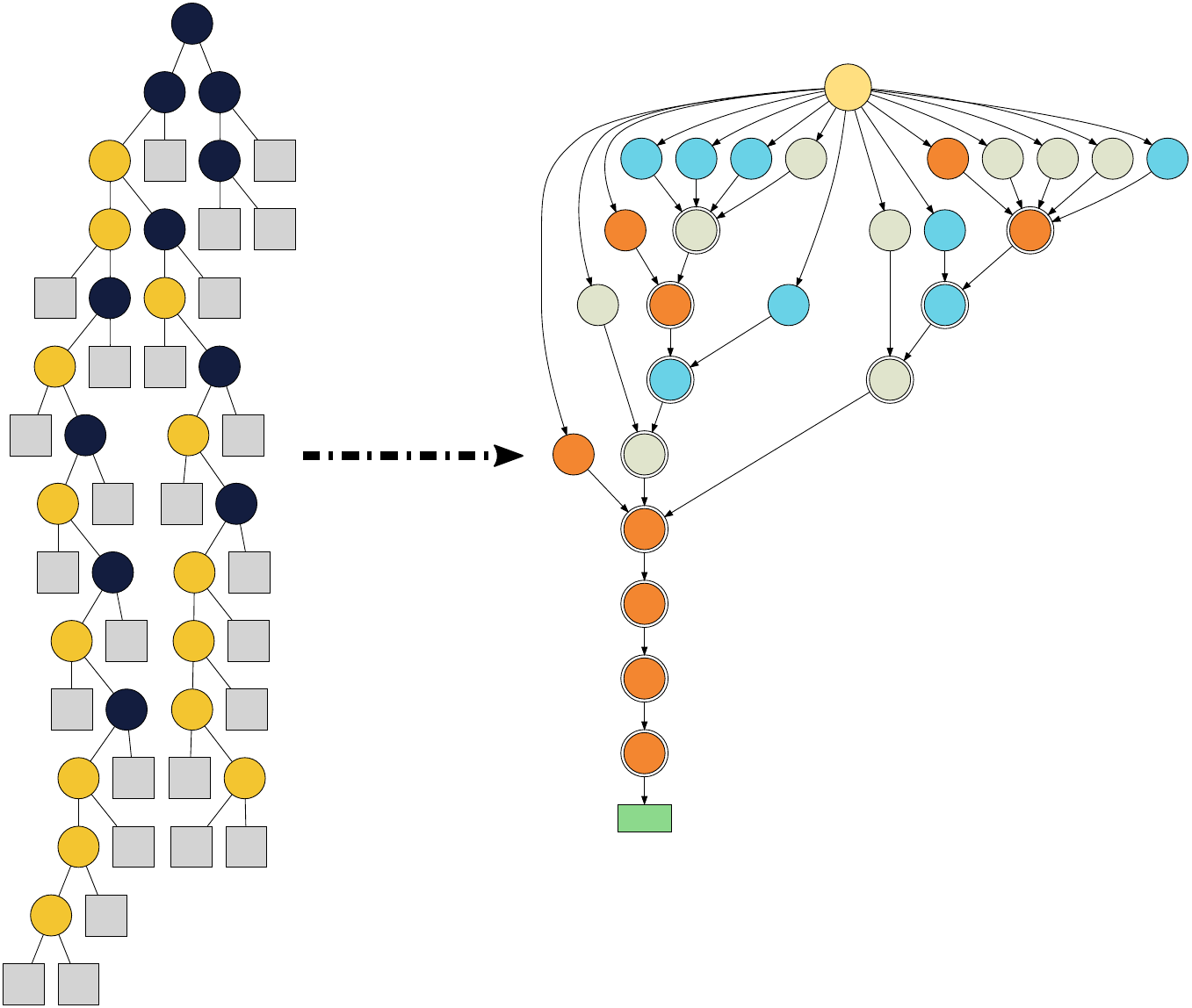}
    }
    \caption{Fittest evolved genotype and phenotype.}
    \label{fig:best_gentoype_phenotype}
\end{figure}

The combined performance of the evolved surrogate models over each generation for 30 runs is presented in Figure \ref{fig:generation_box_plot_all_runs}. It can be observed that most surrogate phenotypes have attained a validation accuracy above 80\% even before the first generation. This indicates that the reduced cellular encoding scheme using the chosen convolutional block design is performant. However, it is also observable that there are still a few phenotypes with low validation accuracy after the evolutionary process has begun.
\vspace{-5mm}
The fittest performing surrogate model was evolved during the 27th generation, as can be seen in Figure \ref{fig:generation_box_plot_all_runs}. The corresponding genotype and phenotype are presented in Figure \ref{fig:best_gentoype_phenotype}. It is of interest to note that the model is both wide and deep - similar to the fittest phenotype found by SurDG-Random. It would appear that the fittest performing phenotype has built a rich feature representation in the wide part of the network and successfully extracted hierarchical relationships in the deep part of the network. In Figure \ref{fig:best_gentoype_phenotype}, the GP tree structure (genotype) shows that two SEQ operations were executed first, creating an initial network of six convolutional layers. 
It may be that in order for a potentially wide network to survive the first few generations, its genealogy may need to start with models that are deep first and then spread out wide in later generations.

\subsubsection{Genealogy Analysis}
\begin{figure}[H]
  \centering
  \scalebox{0.35}{
        \includegraphics[width=1.0\textwidth]{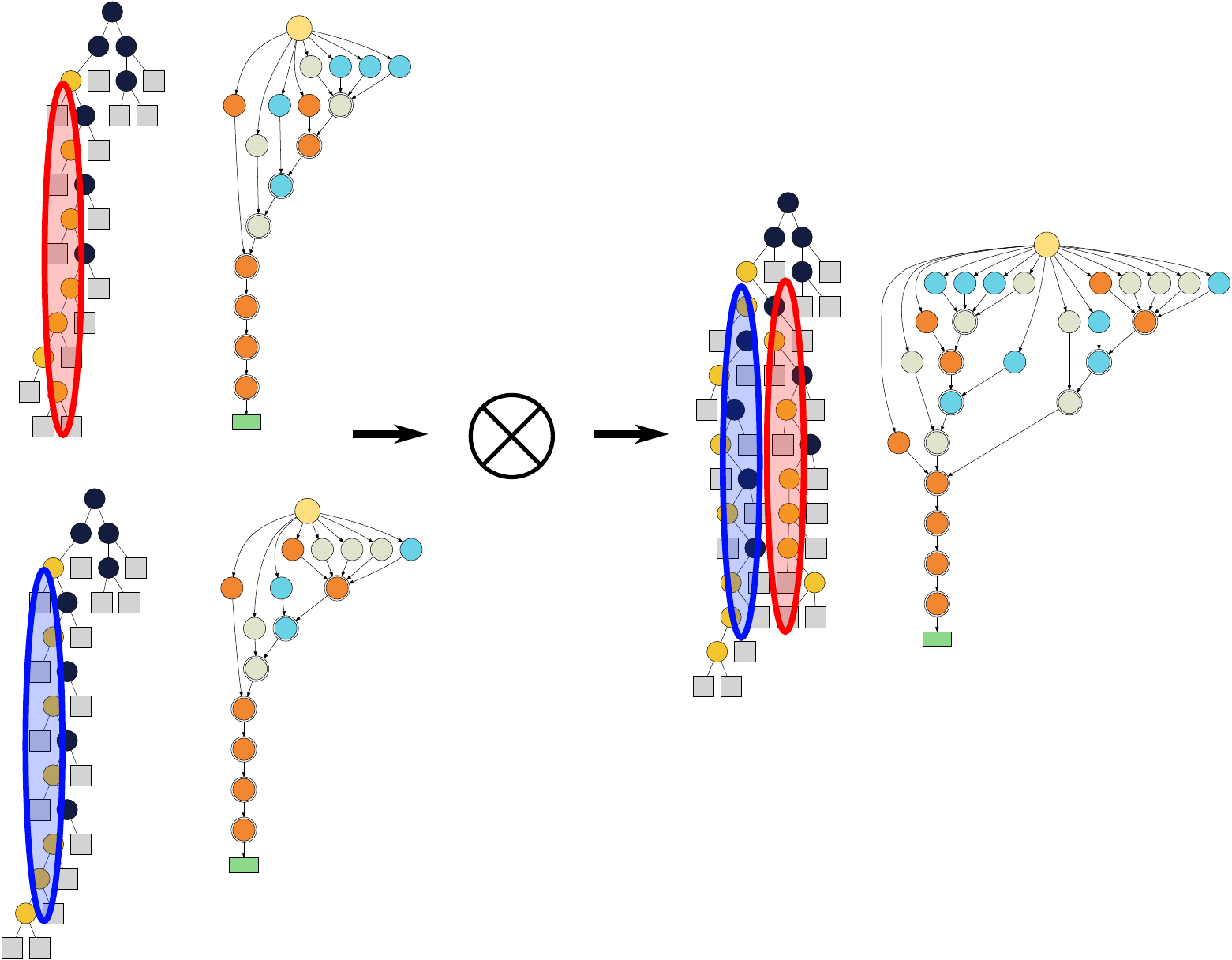}
    }
    \caption{Crossover operation that produced fittest phenotype.}
    \label{fig:crossover_history}
\end{figure}

Analysis of the fittest phenotype's genealogy shows that a crossover operation generated its genotype. This crossover operation is presented in figure \ref{fig:crossover_history}. The components involved in the crossover operation are highlighted in blue and red. Note that both parents are deep networks, and both are wide at the early stages of the topology. The resulting phenotype is as deep as the parents but wider in the early stages. The width of the child phenotype is effectively a summation of the parent's width. There were no mutation operations over the entire genealogy of the fittest phenotype. The lack of a contributing mutation operator raises the question of how important mutation is in the evolutionary process for this particular encoding and is left for future research.
\hspace{5mm}
\subsection{Results of Full Precision Model}
\begin{figure}[H]
  \centering
  \scalebox{0.52}{
  \hspace{-0.5cm}
        \includegraphics[width=1.0\textwidth]{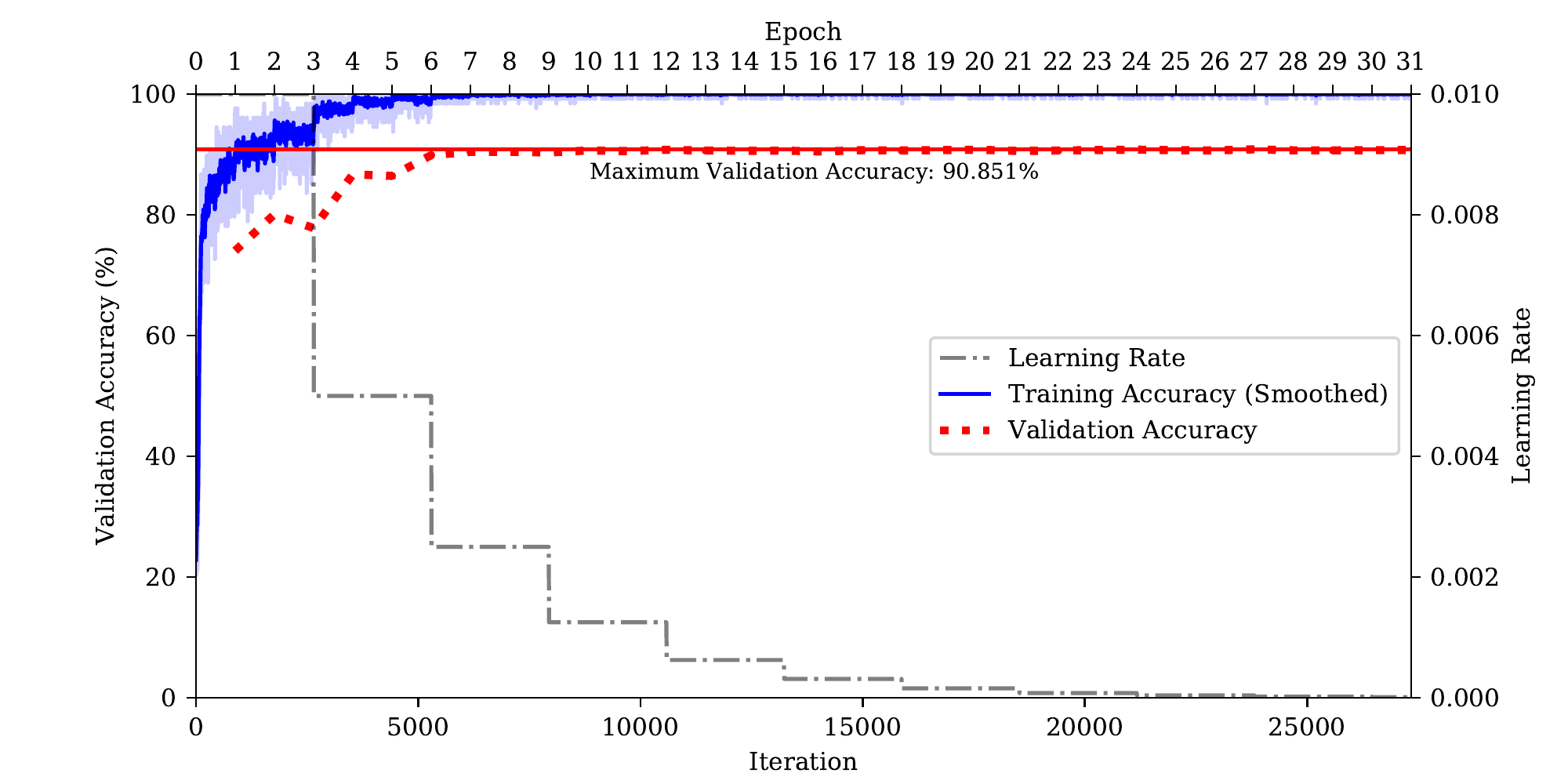}
        }
    \caption{AG's News: Training of high resolution phenotype.}
    \label{fig:training_response_32_bit_phenotype}
\end{figure}

The training and validation history of the full resolution version of the fittest evolved phenotype is presented in Figure \ref{fig:training_response_32_bit_phenotype}. 
The network converged before the seventh epoch, reflecting the successful application of evolutionary pressure applied by the SurDG-EC algorithm on its population to converge before the tenth epoch. The validation accuracy declined after the second epoch until the learning rate was halved at the third epoch, after which the validation accuracy began improving again. The validation accuracy continued improving between epoch five and six but plateaued after the learning rate was halved again. This adverse effect may indicate that the learning rate was too low to escape a local optima, thereby stalling any future accuracy improvements. The importance of dynamically adjusting the learning rate during training is left for future research.
\subsection{Results on AG's News test dataset}
The test performance of the full resolution phenotype and peer competitors is listed in Table \ref{tab:modelResults}. The SurDG-EC algorithm evolved a surrogate phenotype that when trained as a full resolution phenotype, outperformed six of the competitors, including all word-level models. The evolved phenotype compared favourably with the current state-of-the-art VDCNN-29-Kmax model. The fittest evolved phenotype contained roughly half the number of parameters found in the VDCNN-29-Kmax model. It should be noted that the VDCNN-29-Kmax parameter count includes the trainable parameters of its fully connected layers. Interestingly, SurDG-Random located a phenotype that outperformed four expert-designed peer competitor models with a comparable number of parameters.
\begin{table}[H]
    \centering
    \scalebox{0.9}{
    \begin{tabular}{lrr}
        \hline
        Model or Algorithm  & Test Accuracy (\%) & Params. (Millions)\\
        \hline
        word-LSTM (w2v) & 86.06 & -\\
        Large word-CNN (w2v)       & 90.08 & -\\
        Small word-CNN (w2v)       & 88.65 & -\\
        Large Char-CNN \cite{Zhang2015} & 87.18 & \textasciitilde15\\
        Small Char-CNN \cite{Zhang2015} &  \textcolor{red}{\textbf{84.35}} &  \textasciitilde11 \\
        VDCNN-29-Kmax \cite{Conneau2017} & \textcolor{blue}{\textbf{91.27}} & \textasciitilde17 \\
        \textbf{SurDG-Random} (32-bit phenotype)   & 89.11 & \textcolor{red}{\textbf{\textasciitilde18}}\\
        \textbf{SurDG-EC} (32-bit phenotype) & 90.72 & \textcolor{blue}{\textbf{\textasciitilde9}}\\
        \hline
    \end{tabular}
    }
    \caption{AG’s News: Testing accuracy.}
    \label{tab:modelResults}
\end{table}
\subsubsection{Architecture generalisation ability}
To test how well the fittest surrogate phenotypes generalised across other text classification domains, they were retrained as full resolution phenotypes and trained and evaluated across each of the remaining unseen datasets. 
\vspace{0.5cm}
\begin{figure}[H]
\centering
\subfloat[SurDG-Random]{
  \includegraphics[width=0.44\linewidth]{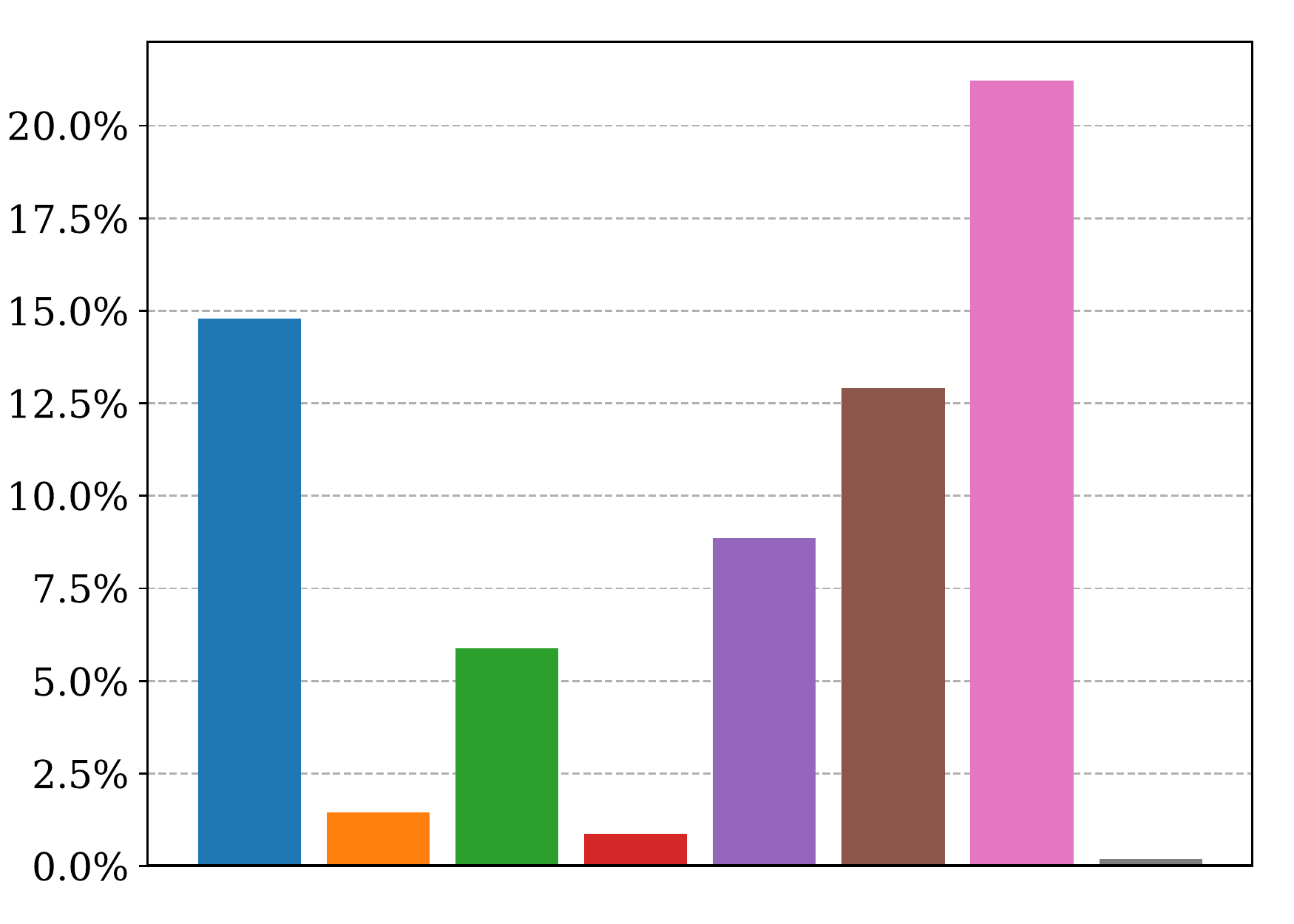}
  }
\label{rp_random_search}
\subfloat[Zhang Small]{
  \includegraphics[width=0.44\linewidth]{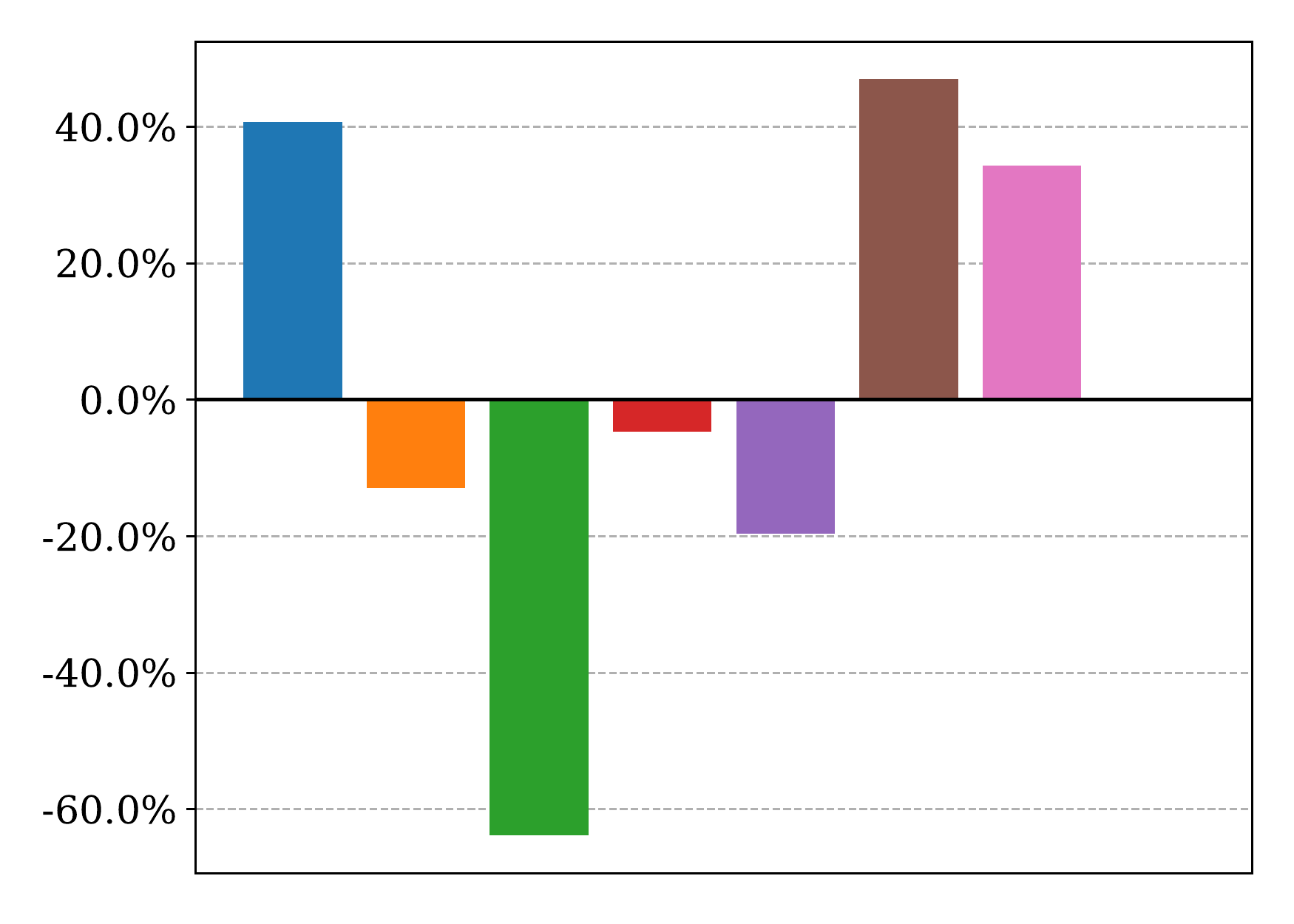}
  }
  \label{rp_zhang_small} \\
\subfloat[Zhang Large]{
  \includegraphics[width=0.44\linewidth]{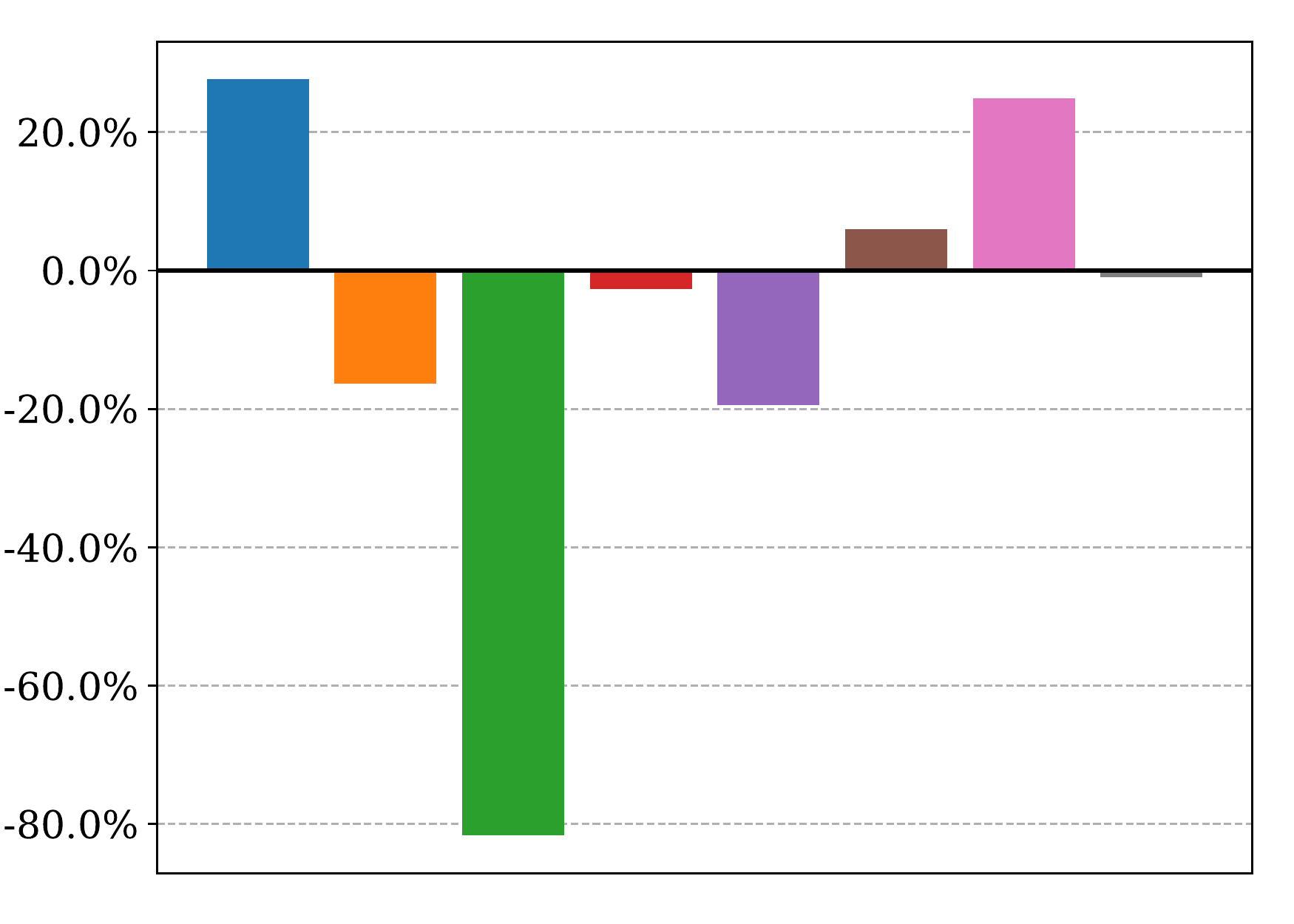}
  }
\label{rp_zhang_large}
\subfloat[VDCNN-29-Kmax]{
  \includegraphics[width=0.44\linewidth]{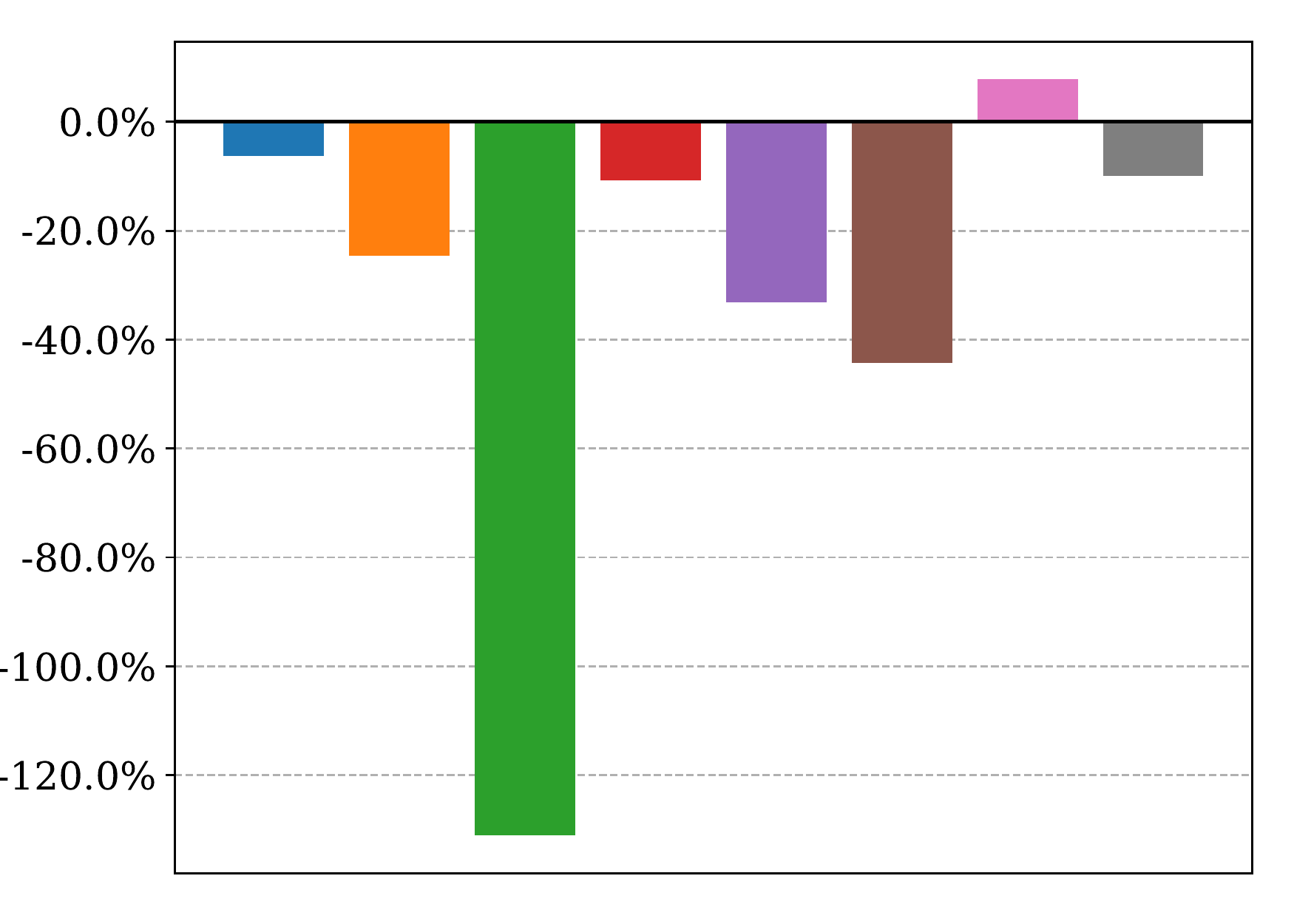}
  }
\label{rp_vdcnn29kmax} \\
\scalebox{1.2}{
\hspace{-8mm}
  \includegraphics[width=1.0\linewidth]{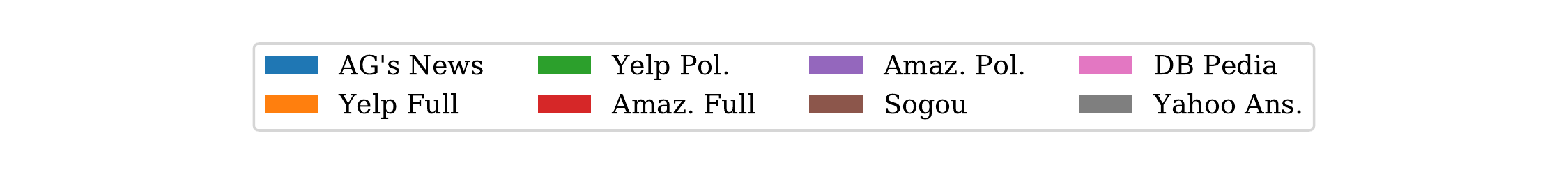}}
\label{rp_legend}
\vspace{-0.5cm}
\caption{SurDG-EC's relative performance.}
\label{fig:relative_Accuracys}
\end{figure}

The relative accuracies are presented in Figure \ref{fig:relative_Accuracys}. It can be seen that the SurDG-EC evolved phenotype outperformed the SurDG-Random located phenotype on all the datasets. It is noted that the SurDG-EC algorithm performed better by only the slightest of margins on the Yahoo Answers dataset.
\newpage
The SurDG-EC phenotype outperformed the remaining competitors on the Sogou dataset and all peer competitors on the DP Pedia dataset. Comparable results were attained on the Yahoo Answers dataset, slightly less so against VDCNN-29-Kmax. The SurDG-EC phenotype significantly under-performed the remaining competitors on the Yelp and Amazon datasets. It should be remembered that the SurDG-EC phenotype was evolved using only 25\% of the smallest dataset. It is impressive that the evolved phenotype could compete favourably on at least three of the unseen datasets. DB Pedia and AG's News are both datasets containing curated articles, and this may explain the ability of the phenotype to generalise so well across the DB Pedia dataset. Sogou news contains romanised Chinese text (pinyin) which is different from the text contained in AG's News. The SurDG-EC phenotype was still able to generalise enough to be performant on the Sogou dataset. This implies that the SurDG-EC algorithm has generalised reasonably well over some unseen datasets. The Yahoo Answers dataset is large, and this appears not to have hindered the performance of SurDG-EC phenotypes. The SurDG-EC phenotype has not generalised well over the Yelp dataset.
\subsection{Further analysis}
\begin{figure}[H]
\centering
\subfloat[Cell-to-depth ratio vs accuracy]{
  \includegraphics[width=0.5\linewidth]{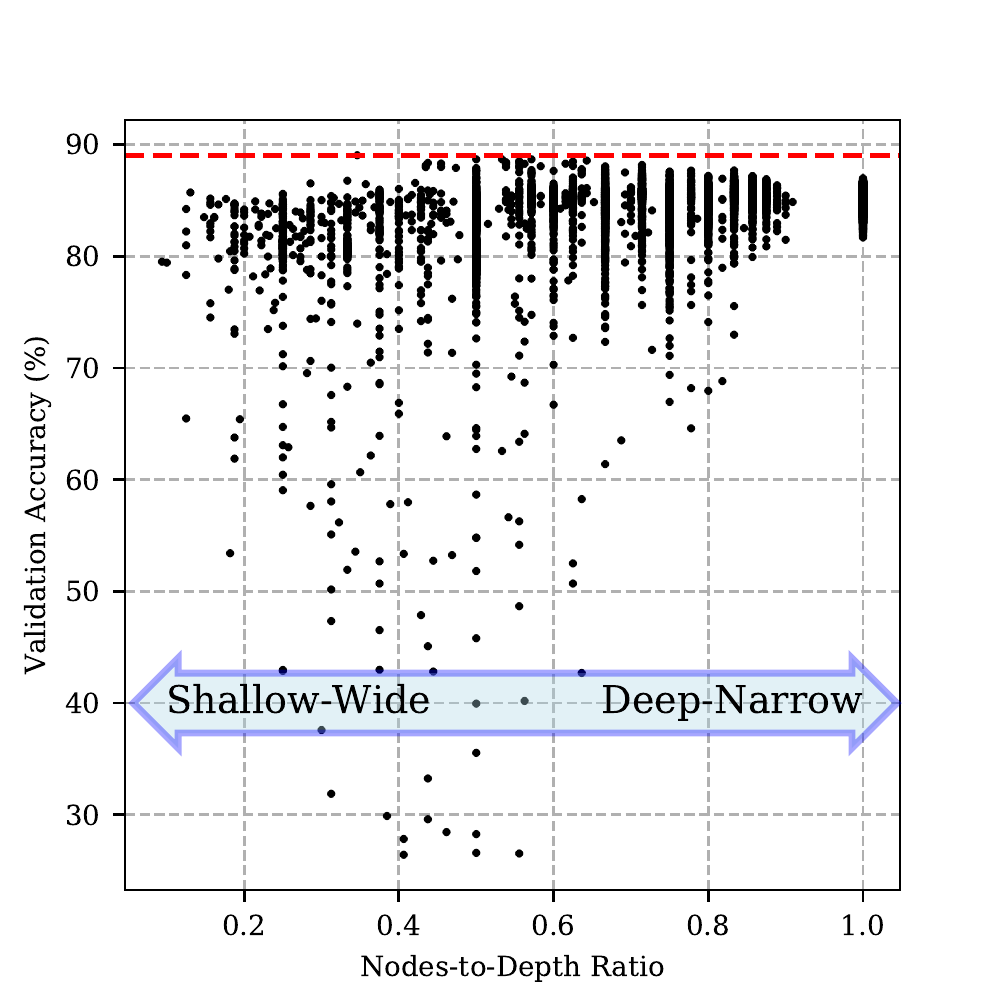}
  \label{fig:nodestodepth_ratio_vs_accuracy}
  }
\subfloat[Path density vs accuracy]{
\hspace{-7mm}
  \includegraphics[width=0.5\linewidth]{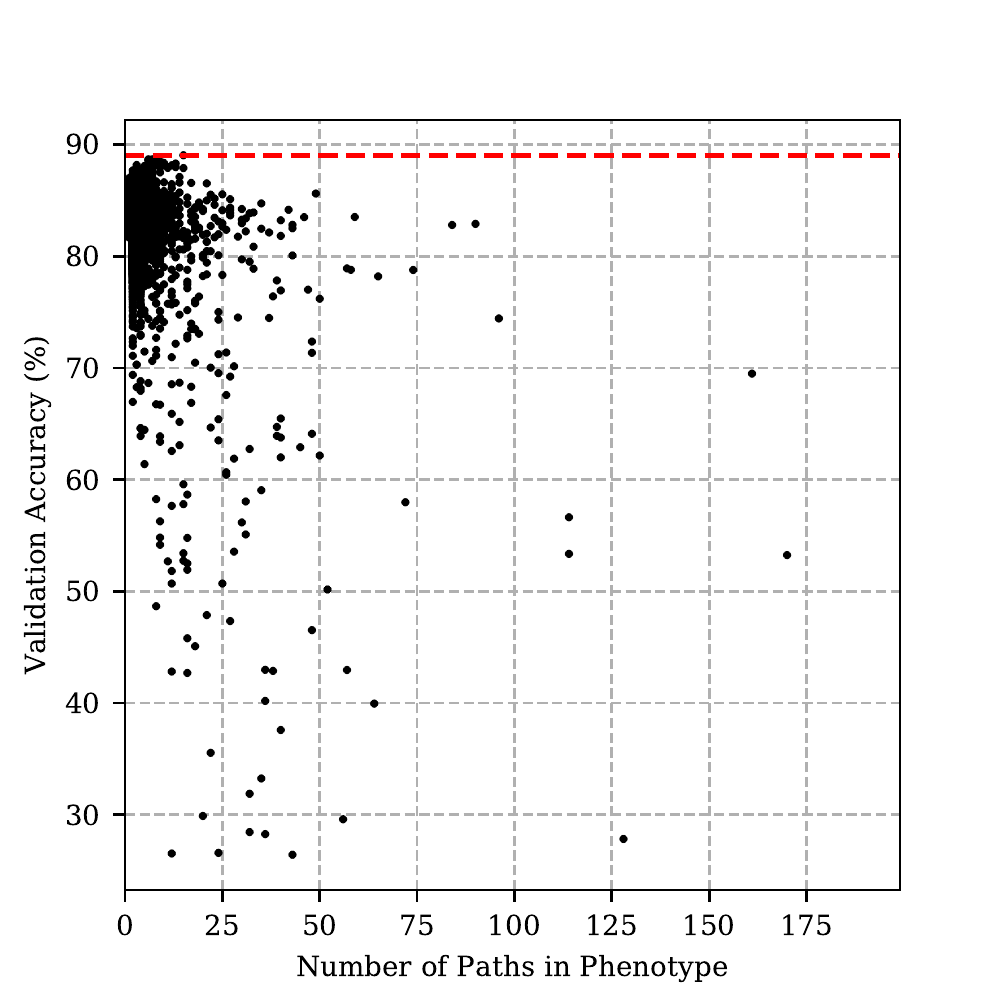}
  \label{fig:paths_density_vs_accuracy}
  }\\
\subfloat[Accuracy vs parameter count]{
  \includegraphics[width=0.5\linewidth]{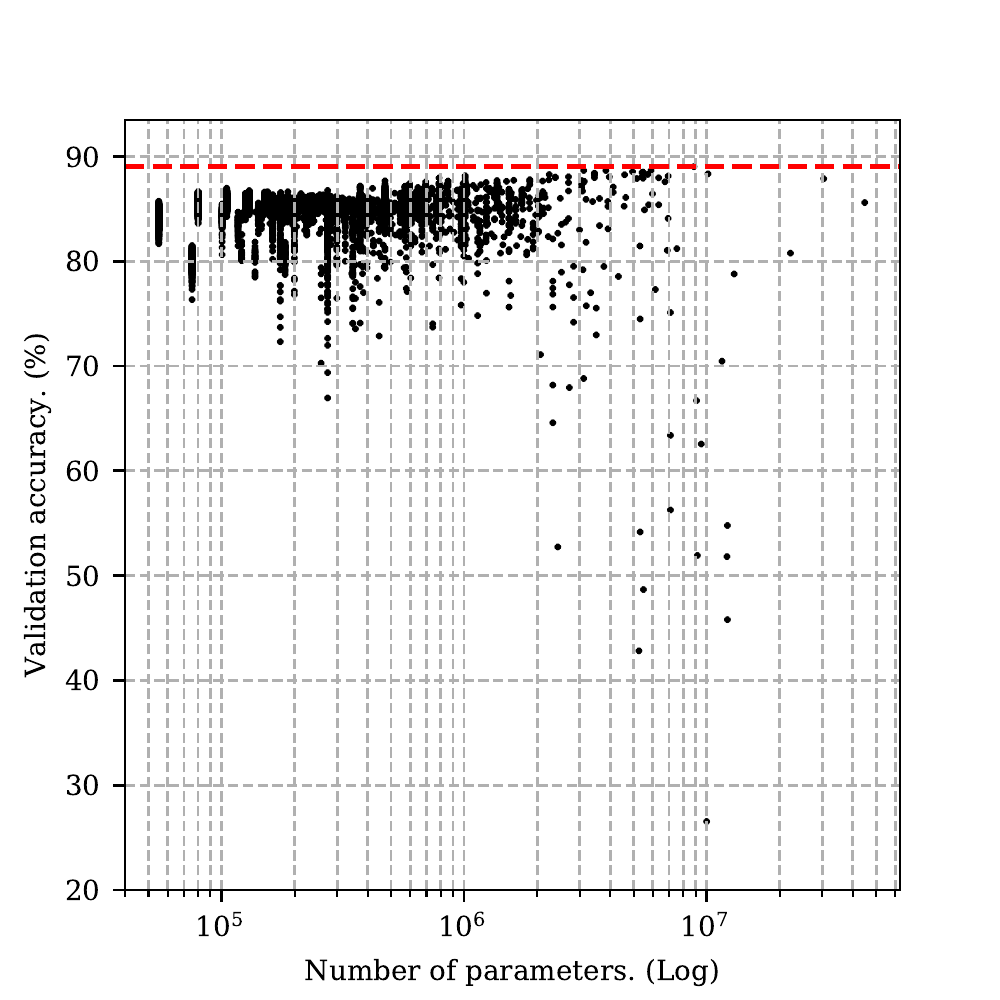}
  \label{fig:params_count_vs_accuracy}
  }
  \hspace{-7mm}
\subfloat[Depth vs accuracy]{
  \includegraphics[width=0.5\linewidth]{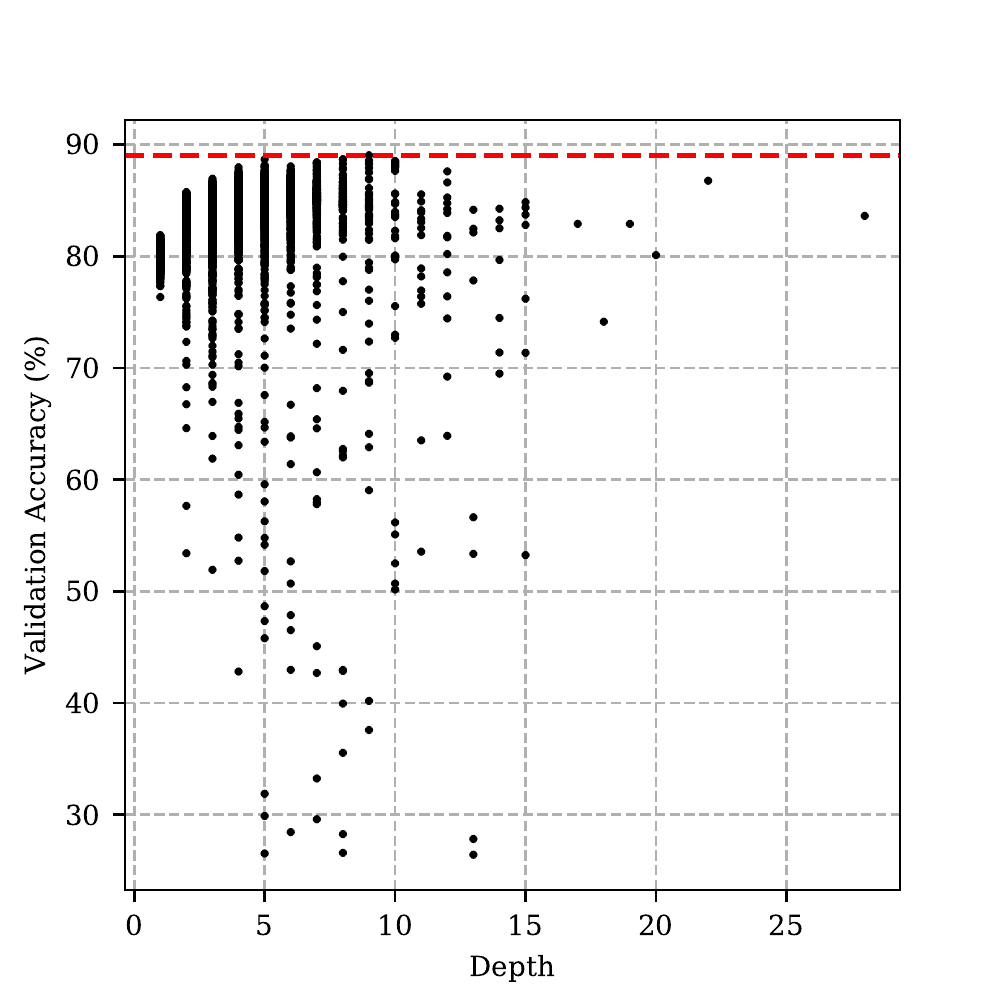}
  \label{fig:depth_vs_accuracy}
  }
  \caption{Surrogate phenotypes analysis.}
  \label{fig:analysis}
\end{figure}

Further analysis was conducted across all distinct surrogate phenotypes evaluated during both SurDG-Random and the evolutionary process. Visualisations of the analysis conducted are presented in Figure \ref{fig:analysis}. Each visualisation represents 14,848 distinct surrogate phenotypes. Four metrics are proposed to aid in determining what properties of a networks architecture contribute to improved or reduced classification accuracy. The proposed metrics are:
\begin{enumerate}
    \item \textbf{Cell-to-depth ratio} is a measure between the number of cells in a phenotype divided by the maximum path length of cells from the input layer to the output layer. A cell-to-depth ratio of 1 implies that all cells are located on the same path. This means the phenotype would have a width of 1. A value approaching zero would imply most cells are on their own path, thus implying the network is wide with a depth of 1. A value between 0 and 1 would imply a network with some combination of width and depth.
    \item \textbf{Path density} is a complexity measure of how many paths an input vector would traverse before arriving at the output. The more paths, the more complex the phenotype is.
    \item \textbf{Trainable Parameters count} is a complexity measure that is simply the number of trainable parameters in the phenotype. A higher value implies a more complex phenotype. 
    \item \textbf{Depth} is a measure that reflects the longest path of sequential cells in a phenotype. The larger the value, the deeper the network.
\end{enumerate}
\newpage
Figure \ref{fig:nodestodepth_ratio_vs_accuracy} presents the cell-to-depth ratio of each surrogate phenotype against their corresponding validation accuracies. The most accurate surrogate phenotypes are located in the centre of the figure, implying that phenotypes with a similar measure of width and depth are the most performant. Although the highest accuracies were attained in this region, there are also many phenotypes with lower accuracies located here too. It would appear that although depth may be important, the situation is more complicated than just merely having a wide network. Deep and narrow networks also attained high accuracies, with no accuracies observed below 80\%. These narrow and deep networks seem to be more performant in general, but unable to achieve as high accuracy as simultaneously wide and deep networks. This lack of high accuracy may also be due to the deep networks getting too deep and not having skip links to improve the training performance. Implementing skip links is left for future research.
Figure \ref{fig:paths_density_vs_accuracy} displays the path density measure of each surrogate phenotype against its corresponding validation accuracy. It is observable that most of the accurate phenotypes have a lower path density measure. This observation implies that phenotypes with too much complex branching and paths may be less performant in general. Most performant phenotypes have less than 25 distinct paths between the input and output.
The parameter count measure is presented in Figure \ref{fig:params_count_vs_accuracy}. It would appear that an increase in the trainable parameter count results in an increase in validation accuracy only up to a certain number of parameters. Beyond this amount, the validation accuracy begins to degrade. This observation is consistent with the findings in \cite{Conneau2017}.
The depth measure is presented in Figure \ref{fig:depth_vs_accuracy}. There is clear evidence that as the depth of phenotypes increase, so does the validation accuracy. After a depth of approximately ten cells, the validation accuracy degrades. It should be mentioned that ten cells represent twenty convolutional layers as defined in \cite{Conneau2017}. The VDCNN-29-Kmax model is 29 layers deep and consists of the same convolutional blocks that constitute a cell in a SurDG-EC evolved phenotype. VDCNN-29-Kmax, however, has skip links in its architecture which enables the model to be deeper than the phenotypes evolved by SurDG. The conclusion drawn is that depth does improve the accuracy of char-CNN up to a certain depth as evidenced in \cite{Conneau2017} and this work.
\subsection{Phenotype Analysis}
\begin{figure}[H]
\scalebox{0.5}{
\hspace{-5mm}
        \includegraphics[width=1.0\textwidth]{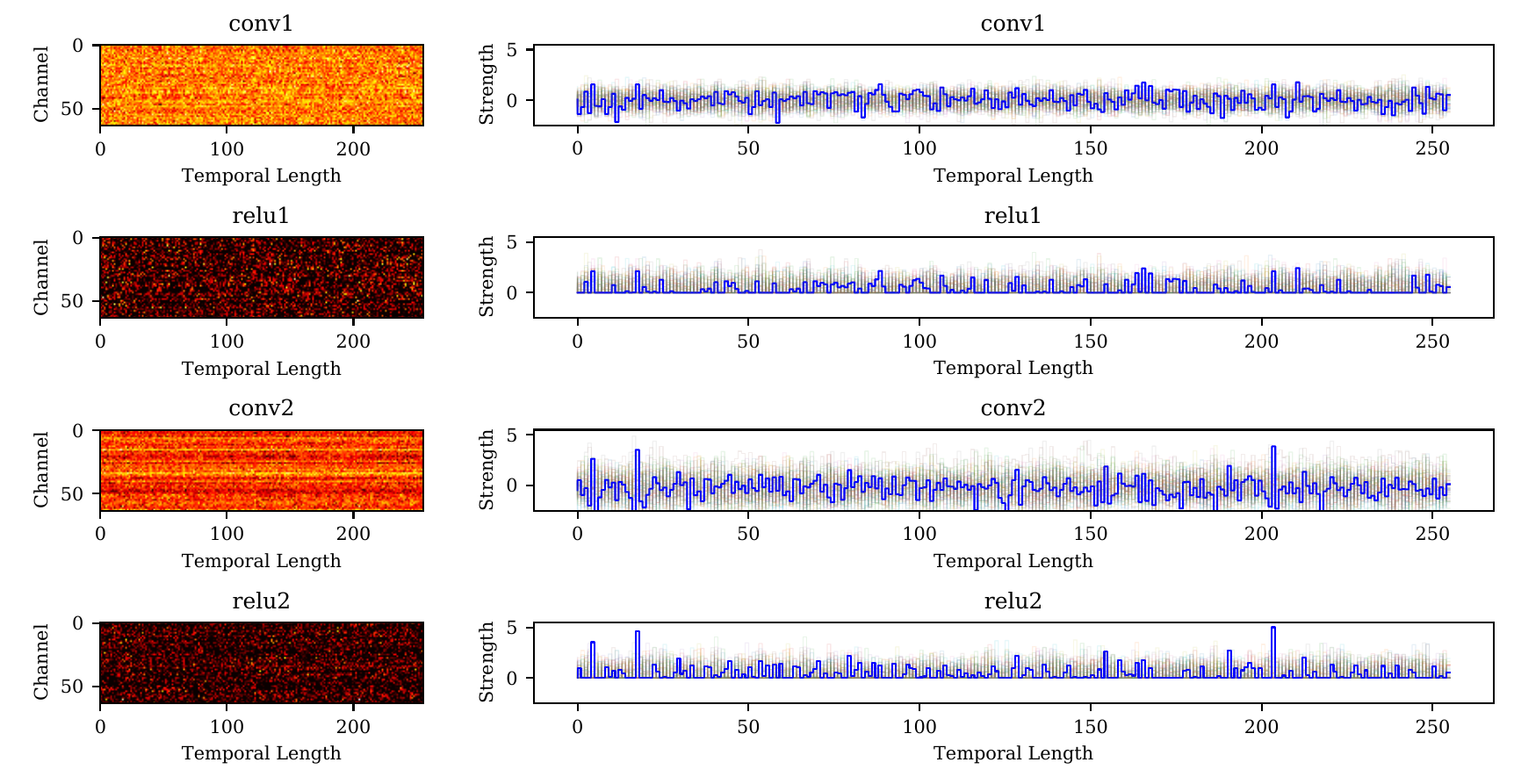}
        }
    \caption{Cellular cell's activations for inference on a single sentence.}
    \label{fig:heatmap_acitvations}
\end{figure}
The feature maps of a cellular cell contained in the fittest full resolution phenotype is presented in Figure \ref{fig:heatmap_acitvations}. A single sentence was sent through the phenotype to capture the activations during inference. The right-hand side of the figure contains a view of the signal produced in channel 60. The channel was selected arbitrarily. There has been little to no research in visualising what takes place within the activations in char-CNNs during inference. It can be observed that as the signal is convolved over, prominent peaks start appearing - representing neurons being excited at that temporal position. There are three prominent spikes after the final ReLU application. It is interesting to note that other channels display many more neurons activating. The value of a channel that has many excited neurons is questionable. Considering the workings of biological evolution, it would make sense that nature would prefer a more efficient encoding of knowledge using a spare representation, meaning less energy consumed. This would imply that a sparse reaction to a stimulus would be preferred. This raises the interesting question of which of the above channels could be pruned and is left for future research.
\section{Conclusion}
This work proposed an evolutionary deep learning approach to discover performant char-CNN architectures. This goal was achieved through the implementation of a genetic programming-based algorithm (SurDG) coupled with a reduced cellular encoding scheme and the backpropogation algorithm. The SurDG-EC algorithm located, on average, higher accuracy models than those located by SurDG-Random. The fittest evolved phenotype defeated one of the state-of-the-art char-CNN models\cite{Zhang2015} and achieved comparable results to the state-of-the-art VDCNN-29\cite{Conneau2017} architecture. The evolved model also generalised favourably across most unseen datasets. There is clear evidence that width may potentially add to the efficacy of char-CNNs.This does not mean that width will always result in increased accuracy, as also observed in the results. There are many other factors to consider. It is not known how much of the efficacy of the evolved phenotypes are due to increased width or some other unknown variable or combination of variables. There are, however, clear indications that the importance of width should be further researched. The SurDG-EC algorithm also revealed two interesting properties of char-CNNs. Building a rich tapestry of feature representations at the early stages of the network potentially aids in improving the accuracy of the networks as they grow deeper - in turn constructing a hierarchy of relations from this rich feature tapestry. The evolutionary crossover operation also revealed that combing the widths of two phenotypes produced a wider phenotype with greater validation accuracy. This is a further clue that there may be value in making char-CNNs with increased width.

\bibliographystyle{ieeetr}
\bibliography{ref.bib}
\vspace{-10mm}
\end{document}